\documentclass[11pt]{article}
\usepackage[margin=1in]{geometry}
\usepackage[T1]{fontenc}
\usepackage{lmodern}
\usepackage{graphicx}
\usepackage{float}
\graphicspath{{./}}
\usepackage{amsmath,amssymb}
\usepackage{booktabs}
\usepackage[numbers,sort&compress]{natbib}
\usepackage{xcolor}
\usepackage[colorlinks=true,allcolors=blue!55!black]{hyperref}
\usepackage{caption}
\title{Off-Axis, On Purpose:\\[0.2em]
\large Where a Transformer Computes Concepts and Why it Does So}
\author{Mark Oskin\\
Professor\\
School of Computer Science and Engineering\\
University of Washington\\
\texttt{mhoskin@uw.edu}}
\date{August 2026}

\begin{document}
\maketitle

\begin{abstract}
\noindent
A transformer's answer lives on one axis: the direction its unembedding reads. Its intermediate
states largely do not, and that off-axis position is usually treated as an obstacle to
interpretation. We show it is functional. A $12$-layer model computes in two phases. Through the
first, every sublayer writes into a subspace held near-orthogonal to the read-out, attention between
$75$ and $96^\circ$ off it at every depth. Moving attention's values onto the read-out is
$64$--$84\times$ more damaging than a matched random rotation, and the damage runs entirely through
cross-token mixing: the off-axis subspace insulates composition from the vocabulary. Beneath that
phase the frame is itself turning: one rotation, fitted to the token stream and never to the
concepts, carries the whole concept constellation from layer to layer at $R^2 = 0.93$, and the turn
is the antisymmetric part of the layer's linear response. In the second phase the answer arrives
on-axis, late---first becoming the top decoded token at layer $9$ of $12$ on average---and by
addition rather than by turning accumulated content onto the read-out.

Pressing every layer onto the read-out instead, the pressure the early-exit line applies, matches the
baseline on perplexity, LAMBADA and BLiMP while cutting the concept-phase workspace from about
twenty-five effective dimensions to fourteen: a change none of those benchmarks register. The
geometry can also be imposed, though not by asking for it. Prescribing it through the loss is a
lottery: six of eight training seeds collapse, because a model told to null its read-out projection
obeys most cheaply by discarding dimensions. Inserting one fixed rotation at the boundary between
the two phases lands it instead, seven of eight seeds at baseline quality with twice the baseline's
effective dimensionality. A sparse rotation, one the surrounding weights can absorb without changing
the function, converges on all nine seeds against five of nine for ordinary unpressured training, and
does so with half the dense device's effective dimensionality, close to the baseline's own. Which rotation is
immaterial: twenty-five runs across thirteen distinct ones reach the same quality, and two baselines
from different seeds hold their concepts in near-orthogonal frames while agreeing on their
read-outs. That freedom is usable. A basis drawn at random and prescribed before training is adopted
across the concept phase, with downstream quality unchanged.
\end{abstract}

\section{Introduction}
\label{sec:intro}

A transformer commits its answer to a single direction. The unembedding reads the final residual
along the axis of the token it is about to predict, and everything the model has computed must
eventually arrive there. Decode the intermediate layers along that same axis and most of the network
looks empty: the logit lens \citep{nostalgebraist2020} resolves the last few layers and
little before them, and the tuned lens \citep{belrose2023} repairs the reading by learning a
per-layer translator, which concedes that the intermediate content is somewhere the read-out axis
does not point. That leaves two questions the lens literature does not answer, and they are this
paper's subject. Why is the working content off the vocabulary axis? And how does the answer come to
be on it?

The usual response is to treat that as an obstacle, or as a defect to be corrected. Work on the
depth alignment sweep frames the off-axis condition as a structural consequence of residual
connections and proposes to mitigate it \citep{lys2026causalshift}; the early-exit line applies the
opposite pressure directly, training intermediate layers to be decodable through a shared read-out
\citep{elhoushi2024layerskip,schuster2022calm}. Both take for granted that content nearer the
read-out is content better placed.

This paper argues the opposite. The off-axis position is doing a job, and the job is insulation.
Attention composes values across tokens by averaging them. If those values carried vocabulary
predictions, averaging them would blur the predictions together, and the blur would land on the axis
the model is trying to write its answer to. Holding the values orthogonal to the read-out puts the
mixing somewhere the unembedding discards, so composition costs the forming prediction nothing. We
measure the cost of removing that insulation: rotating attention's values onto the read-out is
$64$--$84\times$ more damaging than a matched random rotation of the same magnitude, and the damage
is entirely in the cross-token path.

That insulated phase ends, and how it ends is the second thing we measure. Asked whether the answer
is turned out of the accumulated off-axis content or written fresh, the geometry answers plainly.
Across the commit span a best-fit rotation captures only $R^2_{\mathrm{rot}} = 0.38$ of the change,
$64\%$ of the committed state is content orthogonal to any rotation of what came before, and the
residual norm grows by $1.74\times$, which a rotation cannot do. The answer is added rather than
turned. It also arrives late: the final prediction first becomes and remains the top decoded token
at layer $9.2$ on average, and at $39\%$ of positions only at the last layer. A $12$-layer model
therefore computes in two phases, an off-axis one that composes and an on-axis one that writes, with
a boundary between them that is a real location in the network.

If the geometry is functional, the next question is whether it can be imposed rather than observed.
It can, but only in a particular way. Asking for the geometry through the loss, by driving the
residual to exactly $90^\circ$ off the read-out through the first half of the network and onto it
through the second, is a coin flip: six of eight seeds collapse, because a model told to null its
read-out projection can obey most cheaply by throwing away dimensions. Handing the model a frame
instead of demanding one removes the strain. A single fixed rotation inserted into the residual
stream at the phase boundary takes convergence from two of eight to seven of eight, at baseline
quality, with a concept-phase workspace about twice the baseline's effective dimensionality. The
model was never unable to compute off-axis; it was unable to build the frame and hold the pressure
at once.

What the frame has to be turns out to be almost nothing. A sparse rotation that the network's
normalization can absorb reaches the same quality with a workspace half the size of the dense one, so
the dimensionality of the workspace is not what carries the result. Nor does the particular choice
matter: twenty-five runs across thirteen distinct rotations train to the same model, and two baselines trained from
different seeds, aligned by their read-outs, hold their concepts in frames about $90^\circ$ apart,
indistinguishable from unrelated. The $90^\circ$ frame is a free gauge, fixed neither by the task nor
by the device.

A free gauge is also a gauge that can be prescribed. Prescribing a randomly drawn basis in
advance and asking each concept-phase layer to hold its second moment diagonal in it produces models
that adopt the frame we chose, uniformly across the concept phase and robustly to deflating the
largest directions, with downstream quality unchanged. The off-axis phase is therefore not only a
place where things happen to end up. It is a place we can put them.

\paragraph{Contributions.}
We characterize the two-phase geometry of a 12-layer transformer and give a functional account of it
(Section~\ref{sec:baseline}): attention writes $75$--$96^\circ$ off the read-out at every depth, that position
insulates cross-token composition at a measured $64$--$84\times$, the commit is additive rather than
rotational, and the frame the concept phase computes in turns rigidly with depth for reasons we trace
to the layer's linear response (Section~\ref{sec:baseline-rotation}). We quantify the cost of removing
the off-axis position (Section~\ref{sec:forcing}): forcing every layer on-axis reaches parity by
collapsing the workspace, from about twenty-five effective dimensions to fourteen, and does so
invisibly, since perplexity, LAMBADA and BLiMP all stay inside the baseline range. We show that
prescribing the geometry through the loss alone is unreliable (Section~\ref{sec:step}), that supplying
a frame makes it routine (Section~\ref{sec:device}), and that a cheap absorbable frame does the same
job while converging on every seed we ran, against five of nine for ordinary unpressured training
(Section~\ref{sec:sparse}). Finally we show the frame is
both immaterial to the model and ours to prescribe (Section~\ref{sec:gauge}).

The geometry itself---that the middle sits off the read-out axis and is recovered by a learned
decoder---is established, and we take it as given. What this paper adds is the mechanism that
produces that geometry, what it is \emph{for}, and what can be done with it
(Table~\ref{tab:contrib}).

\begin{table}[H]\centering\small
\begin{tabular}{lll}
\toprule
The concept phase\dots & Established by & This paper \\
\midrule
\dots sits off the read-out axis & \citep{nostalgebraist2020,lys2026causalshift} & (taken as given) \\
\dots is readable with a learned decoder & tuned lens \citep{belrose2023} & our instrument \\
\dots is \emph{added to}, not turned on-axis & --- & \S\ref{sec:baseline-additive} ($R^2_{\mathrm{rot}} = 0.38$, $64\%$ new) \\
\dots is \textbf{functional}: it insulates mixing & --- & \S\ref{sec:baseline-offaxis} ($64$--$84\times$) \\
\dots can be \textbf{handed} its frame, not build one & --- & \S\ref{sec:device}--\ref{sec:sparse} (converged $5/9 \to 9/9$) \\
\dots holds that frame as a \textbf{free gauge} & --- & \S\ref{sec:gauge} ($25$ runs, $13$ rotations) \\
\bottomrule
\end{tabular}
\caption{What is established versus what this paper contributes. That the concept phase sits off the
read-out axis and is read by a learned decoder is established and taken as given. That the model
reaches its answer by turning that content onto the read-out---the natural reading of a change of
basis---does not hold: the commit is additive. No decoder of the residual state can test that
difference, since the state is identical whether the prediction was built additively or turned into
place, which is why it needs the geometry (\S\ref{sec:baseline-additive}). This paper
contributes the additive mechanism, the function the off-axis position serves, the frame-supply that
makes the geometry routine to impose, and the gauge freedom that makes it ours to choose.}
\label{tab:contrib}
\end{table}

\section{Experimental setup}
\label{sec:setup}

This section fixes the models, the training recipe, the measurements, and the reporting conventions used throughout.

Every model in this paper is a decoder-only transformer \citep{vaswani2017attention} of the GPT-2 small class: $12$ layers,
$d_{\mathrm{model}} = 768$, $12$ attention heads of width $64$, a feed-forward inner width of
$3072$, GELU activations, pre-layer-normalization blocks, learned positional embeddings, no biases,
and input and output embeddings tied. The vocabulary is the GPT-2 byte-pair vocabulary at $50{,}257$
tokens and the context length is $2048$. This gives $125{,}143{,}296$ parameters, of which
$40{,}170{,}240$ are embedding (token and positional) and $84{,}934{,}656$ are the attention and
feed-forward blocks. We report the split because the interventions in Sections~\ref{sec:device}
and~\ref{sec:sparse} change the block parameters and leave the embedding untouched, and because at
this scale the embedding is a third of the total.

We train on OpenWebText \citep{gokaslan2019openwebtext}, approximately $8.9$ billion tokens, for one
epoch, with a held-out validation split of $120$ million tokens. Optimization is AdamW at
$\beta = (0.9, 0.95)$, weight decay $0.1$, peak learning rate $3\times10^{-4}$ with $2000$ warmup
steps and a cosine decay to a tenth of peak, batch size $16$ at the full $2048$-token context,
gradient clipping, and \textsc{bf16} mixed precision. One epoch is $272{,}687$ steps. Runs execute on
a single H200 and differ only in the training seed unless stated otherwise. We do not tune a run back to health, for the reason given in Section~\ref{sec:baseline-convergence}: the
convergence rate is one of the quantities we are measuring, and repairing failures by hand would
remove it.

The rotations of Sections~\ref{sec:device}--\ref{sec:gauge} are fixed at initialization from a
per-run seed, never trained, and never dependent on the input. Where a device and a loss term are
used together, the device is applied inside the forward pass and the loss term is added to the
cross-entropy objective with a fixed coefficient held constant through training. Devices add no
parameters.

We report zero-shot LAMBADA accuracy \citep{paperno2016lambada}, BLiMP \citep{warstadt2020blimp},
and validation perplexity. Benchmarks are evaluated with right-padding and an attention mask that
excludes padded positions; scoring several sequences of unequal length in one batch without both is
a silent corruption, and one that changes the ranking between models rather than shifting all of
them together. All benchmark figures in this paper come from a single harness applied uniformly to
every model.

Each quantity is the average over the \emph{converged} runs of a flock, with the range across that
set given in tables, and the convergence rate reported separately as a measurement in its own right
(Section~\ref{sec:baseline-convergence}). Runs that fail to converge are excluded from averages
rather than repaired, and are reported as failures. Because a run can look healthy at an
intermediate checkpoint and be degenerate at the end, we screen every run that enters an average by
its final residual participation ratio as well as by its perplexity; a model whose residual has
collapsed to a handful of directions is not a usable model even when its loss looks ordinary.

A substantial fraction of from-scratch runs fail, and that attrition is characteristic of the setting rather than of our recipe. Large-model training logs report that from-scratch optimization diverges as a matter of course: the PaLM loss spiked roughly twenty times despite gradient clipping and recovered only by restarting from an earlier checkpoint and skipping batches \citep{chowdhery2022palm}, and the OPT-175B log records at least $35$ manual restarts from hardware failures, with lowering the learning rate and resuming from an earlier checkpoint reported separately as the recovery for a diverged loss \citep{zhang2022opt}. At our exact scale the rate has been measured directly: a GPT-2 $124$M model trained on OpenWebText in \textsc{bf16}, the same model and corpus we use, diverges on about one run in ten within five percent of the training budget, which the authors describe as a lower bound \citep{lee2024}, and an entire toolkit, warmup, query-key normalization, z-loss and $\mu$P among them, exists to suppress the effect \citep{wortsman2024}. We do not repair a failed run by hand, and we judge each by a downstream zero-shot metric, so this instability appears in our numbers rather than being engineered away. Where a warm restart was attempted we report it as such and exclude it from the converged averages. We therefore report each quantity as the average over the converged set, give the range across that set in tables, and treat the convergence rate, the fraction of seeds that yield a usable model, as a measurement in its own right. A carefully tuned pipeline fares better, though outliers persist: a systematic seed study reports them at about one run in five at $410$M parameters and roughly one in twenty-five averaged over scales, with the outlier runs completing training rather than failing \citep{vanderwal2025}. That study uses a different scale and recipe, so we offer it as context rather than as a matched comparison; the rate we report reflects the absence of stabilizing machinery.

Three of the measurements in this paper compare a quantity against a null, and in each case the null
is matched rather than generic, because the generic version is uninformative. The insulation
measurement of Section~\ref{sec:baseline-offaxis} compares a rotation toward the read-out against a
random rotation \emph{of the same magnitude}, and separately against the same rotation with
cross-token averaging removed, which is what isolates the mixing as the cause. The rigidity
measurement of Section~\ref{sec:baseline-additive} uses a rotation fitted only to the token stream
and never to the concepts, and reports it against the no-motion alternative rather than against zero.
The frame prescriptions of Section~\ref{sec:gauge-named} are measured against foreign bases drawn the
same way as the prescribed one, not against an arbitrary reference. Angles near $90^\circ$ in high
dimension and fits to small numbers of points are both easy to over-read, and a matched null is what
separates the two.

\subsection{Geometric measurements}
\label{sec:methods}

Three measurements recur throughout the paper, and we define them once here.

\paragraph{Participation ratio.}
To count how many directions a representation uses, we compute the participation ratio of the residual stream at each layer. With $\lambda_i$ the eigenvalues of the centered covariance of the residual activations,
\begin{equation}
\mathrm{PR} \;=\; \frac{\left(\sum_i \lambda_i\right)^2}{\sum_i \lambda_i^2}.
\end{equation}
The participation ratio is the number of equally active dimensions that reproduce the same spectrum: it equals the width $d$ for an isotropic representation and $1$ for a rank-one one. It is a standard measure of effective dimensionality: the inverse participation ratio in physics, and the dimensionality of population activity in neuroscience \citep{litwinkumar2017,gao2017}, and a continuous counterpart of the effective rank \citep{royvetterli2007}.

\paragraph{Orthogonal Procrustes.}
To ask whether the network reaches a later state by turning an earlier one or by adding to it, we use the orthogonal Procrustes fit. Given centered residual states $X$ and $Y$ at the two ends of a span of layers, the rotation minimizing $\lVert XR - Y\rVert_F$ over orthogonal $R$ has the closed form $R = UV^\top$ from the singular value decomposition $X^\top Y = U\Sigma V^\top$ \citep{schonemann1966}. We report $R^2_{\mathrm{rot}}$, the fraction of the layer-to-layer change captured by this best-fit rotation, and the complementary new-content fraction, the part of $Y$ orthogonal to $RX$. As a control, inserting an explicit fixed rotation between two layers of a network drives $R^2_{\mathrm{rot}}$ to $1.00$, confirming that the decomposition registers a rotation when one is present.

\paragraph{Write angle to the read-out.}
Each sublayer: an attention block or a feed-forward block, contributes an additive write to the residual stream. To ask whether a write points toward the eventual prediction, we send it through the final layer normalization and the tied unembedding, the logit lens \citep{nostalgebraist2020,belrose2023}, and measure the angle between the resulting vocabulary-space vector and the model's final logits, averaged over token positions. An angle near $90^\circ$ places the write in a subspace the unembedding discards; an angle near $0^\circ$ aligns it with the answer. That feed-forward writes carry vocabulary-space content at all is established \citep{geva2021,geva2022}; we track the angle of that content across depth and by sublayer, within the residual-stream view of the transformer \citep{elhage2021}.

\section{The GELU baseline}
\label{sec:baseline}

Every measurement in this paper is read against a picture of the untouched network, and this section builds that picture. We train nine models of the GPT-2 small class on OpenWebText under the recipe of Section~\ref{sec:setup} \citep{radford2019gpt2,gokaslan2019openwebtext}, identical but for the random seed, and ask two questions of them: where in the residual stream the network computes, relative to the vocabulary read-out; and how reliably the recipe yields a usable model at all.

One term needs fixing before the measurements make sense, because the rest of the paper turns on it. The unembedding reads the final residual along one direction for each token in the vocabulary. A residual component lying along those directions is a claim about which token comes next; a component orthogonal to all of them is invisible to the unembedding and contributes nothing to the prediction. The read-out therefore supplies a privileged set of directions in an otherwise arbitrary space, and every angle in this paper is measured against it. To say the network computes \emph{off-axis} is to say it holds its intermediate content in directions the unembedding throws away.

\begin{figure}[H]
\centering
\includegraphics[width=0.92\textwidth]{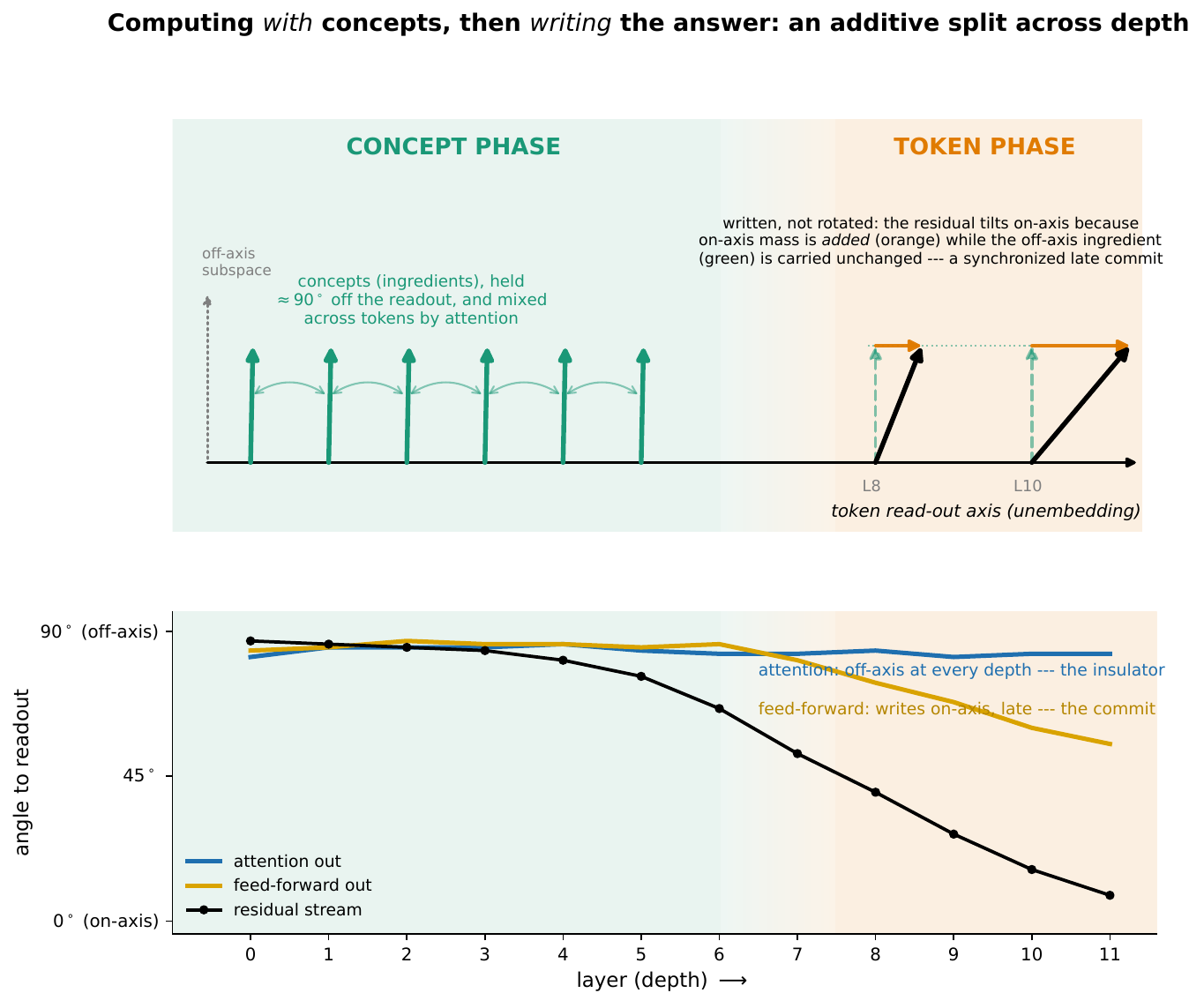}
\caption{Computing with concepts, then writing the answer. Through the concept phase the residual carries ingredients held $\approx\!90^\circ$ off the read-out and mixed across tokens by attention; at the commit ($\approx$ layer~9, with $84\%$ of positions resolving over layers~8--11) a fresh on-axis component is added while the off-axis ingredient is carried forward, so the residual leans on-axis by addition. The lower panel shows the measured write angles across depth: attention stays off-axis at every layer, the feed-forward path swings toward the read-out late, and the residual stream reaches it.}
\label{fig:schematic}
\end{figure}

Figure~\ref{fig:schematic} states the geometric answer in outline. Through the body of the network the computation lives in a subspace held roughly orthogonal to the vocabulary read-out, attention mixes concept ingredients across token positions there, never turning them toward the output axis, and only in the last few layers does the feed-forward path write the prediction onto that axis. The answer is assembled at the read-out late; it is not carried there by turning the middle.

\subsection{Concepts are computed off the read-out}
\label{sec:baseline-offaxis}

\begin{figure}[H]
\centering
\includegraphics[width=0.98\textwidth]{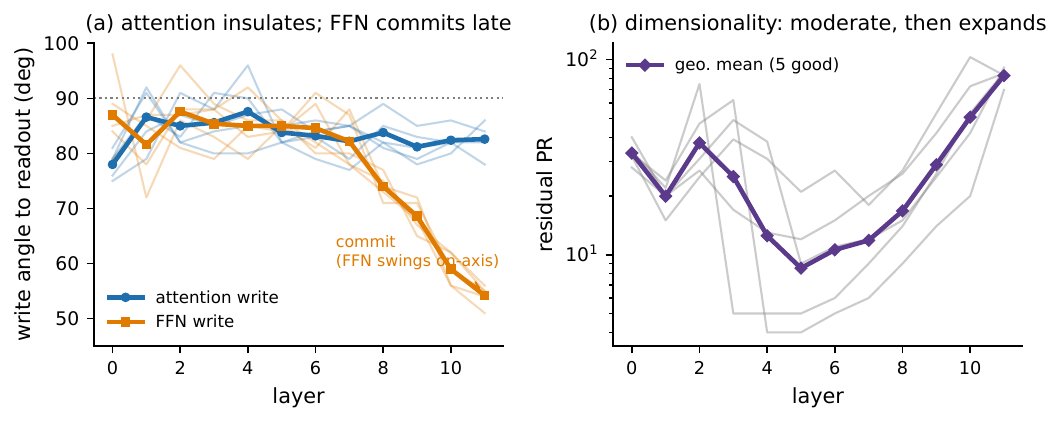}
\caption{Per-layer geometry of the five converged baselines (thin lines individual runs, bold their mean). (a)~The angle of each sublayer's write to the read-out: attention (blue) stays $\approx\!83^\circ$ off-axis at every depth, the insulator, while the feed-forward write (orange) swings toward the read-out over the last four layers, the commit. (b)~The residual participation ratio: moderate through the body of the network and expanding toward the output.}
\label{fig:baseline-geom}
\end{figure}

Figure~\ref{fig:baseline-geom} shows the two geometric measurements across depth for the five converged models, and both patterns hold across seeds. The attention write sits $75$--$96^\circ$ from the final logits at every one of the twelve layers: attention never turns toward the vocabulary. The feed-forward write holds near $85^\circ$ through the body of the network and swings toward the read-out over the last four layers, reaching about $55^\circ$ at the output. The residual participation ratio (panel~b) is moderate through the middle of the network, dropping to roughly ten, and expands sharply toward the output, to seventy or more at the final layer. The network computes in a subspace held roughly orthogonal to the vocabulary throughout its depth, and only in the final layers does the feed-forward path turn content toward the read-out to form a prediction.

The reason for this off-axis organization is in what attention does. Attention mixes across token positions: every position's output is a weighted average of the sequence's values. Were that averaging to run on the read-out axis it would blur the prediction, smearing one position's next-token distribution into its neighbours'. Held orthogonal to the read-out, the same mixing moves information between positions in a subspace the unembedding discards, and the prediction is left untouched. The off-axis workspace is therefore protective: it insulates the read-out from the blur that cross-token averaging would otherwise inflict. This is why attention's writes never turn toward the vocabulary (Figure~\ref{fig:baseline-geom}a), and why, in the attribution of Section~\ref{sec:baseline-additive}, attention contributes essentially nothing to the answer direction. Attention is the mixer, and the vocabulary axis is kept clear of it until the feed-forward path writes the prediction on-axis at the end.

\begin{figure}[H]
\centering
\includegraphics[width=0.62\textwidth]{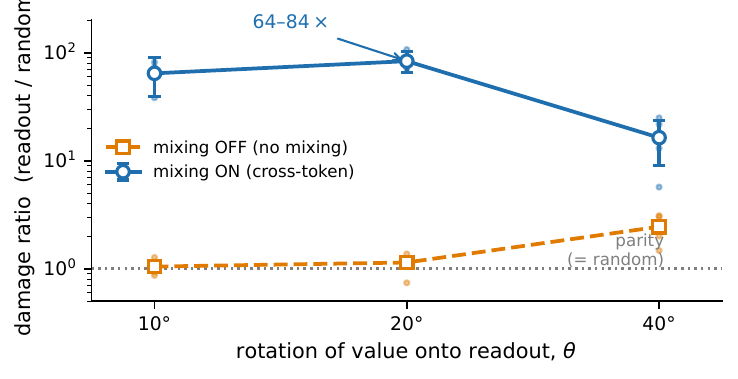}
\caption{Off-axis positioning is functional insulation for cross-token mixing (five converged baselines, each against four draws of the control rotation). Rotating attention values toward the read-out and letting attention mix them is $64\times$ more damaging to the next-token loss than a matched random rotation of the same magnitude at a $10^\circ$ rotation, and $84\times$ at $20^\circ$ (blue); removing the cross-token averaging, each position keeping its own rotated value, collapses the asymmetry to $\sim\!1\times$ (orange). The damage is entirely in the mix.}
\label{fig:insulation}
\end{figure}

This is not only an argument from geometry; the cost of violating it is measurable (Figure~\ref{fig:insulation}). On a trained baseline we rotate each attention head's value vectors toward the read-out by a small angle and let attention mix them as usual. Averaged over the five converged models, each against four draws of a random rotation of the same
magnitude, the damage this does to the next-token loss is $64\times$ that of the random control at a
$10^\circ$ rotation (range $38$--$82$ across models) and $84\times$ at $20^\circ$ (range
$71$--$107$). A matched control isolates the cause: apply the identical rotation but strip the
cross-token averaging, each position keeping its own rotated value, and the asymmetry collapses to
$1.05\times$ and $1.14\times$ at those same two angles. Read-out-aligned values are no worse than
randomly rotated ones until they are \emph{mixed}; the damage is entirely in the mix. The effect is
one of small perturbations and we report it as such: at $40^\circ$ the ratio falls to $16\times$,
because a rotation that large degrades the model whichever direction it points, leaving the control
without a floor to measure against. The off-axis workspace is therefore functional insulation rather than an incidental preference: the network holds its values near $90^\circ$ off the read-out precisely to keep attention's averaging from blurring the prediction.

That argument is about attention, and the geometry does not let us stop there. Through the body of the network the feed-forward write is just as far off the read-out as attention's, $84.5^\circ$ against $84.9^\circ$ averaged over layers~1--7, the per-layer means of both spanning $82$--$88^\circ$; the two separate only at the commit, where attention holds near $82^\circ$ while the feed-forward path swings to $64^\circ$ (Figure~\ref{fig:baseline-geom}a). A feed-forward layer mixes nothing across token positions. It has no blur to protect against, so its off-axis position cannot be insulation in the sense measured above.

The two facts fit together without a second mechanism. The off-axis subspace is a shared workspace, and insulation is the reason that workspace has to be off-axis \emph{at all}: were it on the read-out, attention's averaging would blur the prediction at the cost the rotation experiment measures. Once it sits off-axis, everything that writes into it writes off-axis, the feed-forward path included, because that is where the content it reads and adds to already lives. The insulation result is scoped accordingly. It is a claim about attention's values and about mixing, which is what the matched control isolated, and it explains where the workspace sits rather than the direction of every write into it.

\subsection{The answer is written to the read-out, rather than turned from the concept space}
\label{sec:baseline-additive}

\begin{figure}[H]
\centering
\includegraphics[width=0.82\textwidth]{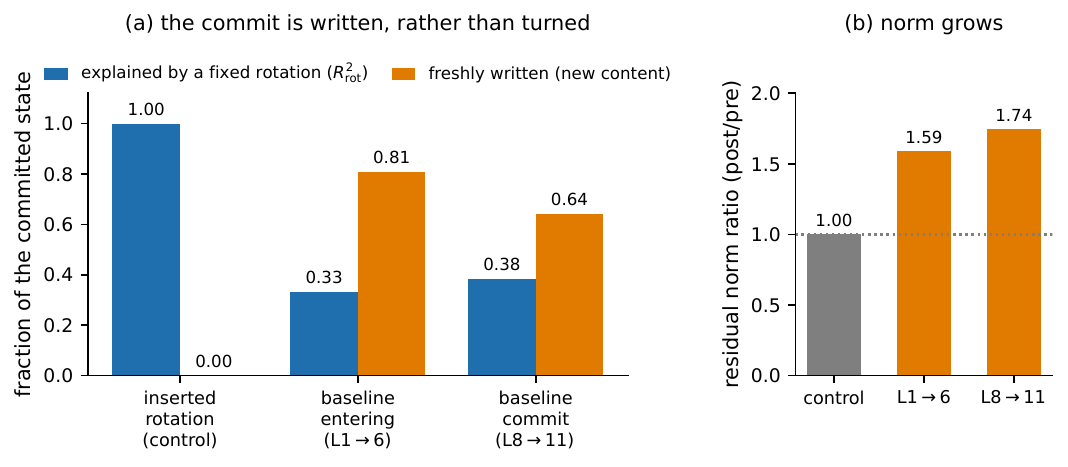}
\caption{The commit is written, not rotated (mean over the five converged runs). (a)~A best-fit rotation explains only a minority of each transition ($R^2_{\mathrm{rot}}$, blue), while most of the resulting state is freshly written content (orange); an explicitly inserted rotation is recovered at $R^2_{\mathrm{rot}}=1.00$ as a control. (b)~The residual norm grows across the same spans, as additive writing predicts.}
\label{fig:commit-additive}
\end{figure}

\begin{figure}[H]
\centering
\includegraphics[width=0.62\textwidth]{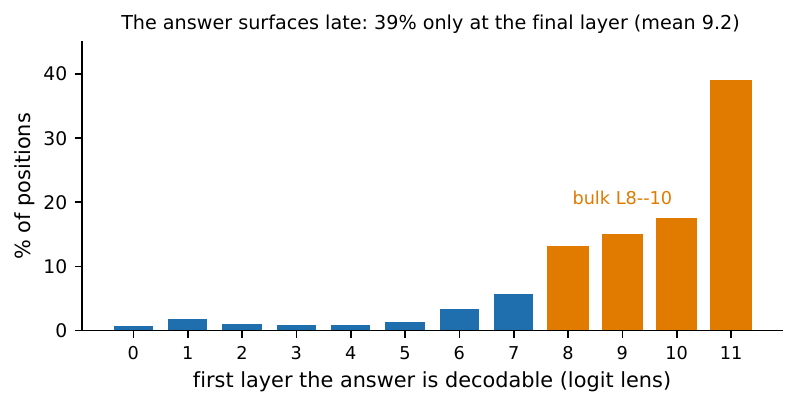}
\caption{The prediction surfaces late (mean over the five converged runs). For each token position, the first layer at which the logit-lens top token matches and holds the final prediction; the mass sits in layers~8--11, with $39\%$ of positions resolving only at the final layer and a mean commit layer of $9.2$.}
\label{fig:commit-layer}
\end{figure}

Is the on-axis prediction a turning of the off-axis body representation, or is it written fresh? The orthogonal Procrustes decomposition answers directly (Figure~\ref{fig:commit-additive}). Across the commit span (layers~8--11) a single best-fit rotation captures $R^2_{\mathrm{rot}} = 0.38$ of the change, while $64\%$ of the committed state is content orthogonal to any rotation of the input. The residual norm grows by $1.74\times$ over the same span; a rotation preserves norm, whereas additive writing grows it. The two mechanisms are distinct: the off-axis representation does move as a near-rigid body from layer to layer, but that turn stays within the off-axis subspace and never delivers the answer, which arrives as a separate on-axis write. We measure the rigidity with a rotation that is never fitted to the concepts: the polar factor of the residual cross-covariance, determined from the token stream alone ($N\!\approx\!250{,}000$ states in $768$ dimensions, so the estimate is fully determined). That single rotation carries the concept constellation from one layer to the next at $R^2 = 0.93$, against $0.76$ for leaving the constellation unchanged, and the figure holds between $0.933$ and $0.938$ across every model we measure. Consistently, the prediction surfaces late: decoded through the logit lens, the layer at which the final token first becomes and remains the top choice averages $9.2$, and $39\%$ of positions resolve only at the final layer (Figure~\ref{fig:commit-layer}). The transition-wise Procrustes split into a rigid rotation and a non-rigid remainder is the instrument of \citet{bhattacharya2026residual}, who apply it across depth in six instruction-tuned models and report the non-rigid part peaking at the final transition. Their measurements are layer-relative and do not reference the unembedding; what the fixed read-out adds here is which part of the transition carries the answer on-axis.

The same measurement on the other boundary says the model enters the workspace the way it later leaves it. Across layers~1--6, where the residual moves \emph{off} the read-out rather than onto it, a best-fit rotation captures $R^2_{\mathrm{rot}} = 0.33$ \;[$0.26$--$0.39$], $81\%$ of the resulting state is fresh content, and the norm grows $1.59\times$. Going off-axis is written rather than turned, and marginally more so than the commit.

If any content were carried by clean rotation, it should show up somewhere as a set of positions that turn by roughly $90^\circ$ while adding little. There is no such set. Fitting a separate rotation for each of the $44$ token types with enough positions to constrain one, over $704{,}512$ positions in each of the five baselines, the strongest candidates are function words and punctuation, tokens whose own embedding is swamped by context and which therefore sit furthest off-axis. Even these reach only $R^2_{\mathrm{rot}} = 0.20$ on average, and $0.50$ for the single best basket in any run, at a mean angle of $82^\circ$ with $83\%$ of the content freshly written. They travel a long way off-axis and they get there additively.

Rotation is therefore not an operation this network performs, in either direction and at no depth we can find it. That is worth establishing before Section~\ref{sec:device} inserts one deliberately: the device is not amplifying a motion the model already makes, it supplies one the model has no native form of.

\begin{figure}[H]
\centering
\includegraphics[width=0.98\textwidth]{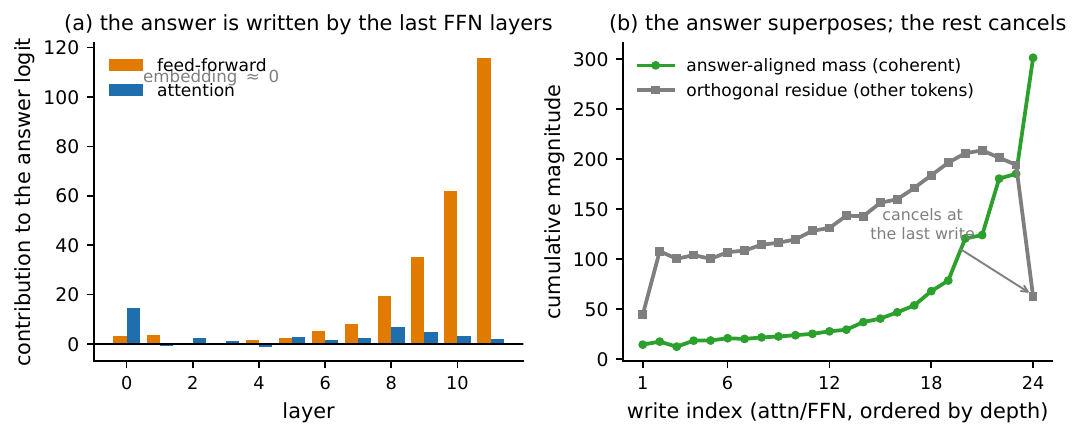}
\caption{How the answer is assembled (direct logit attribution with a shared final normalization, so the sublayer writes sum to the logits exactly; median over the five converged runs). (a)~Each sublayer's contribution to the final-answer logit: the last feed-forward layers write almost all of it, the final layer about a third, while attention and the token embedding contribute nothing. (b)~The answer-aligned mass accumulates coherently across the writes while the orthogonal ``other-token'' residue rises and then cancels at the final write, so the prediction forms by superposition.}
\label{fig:dla}
\end{figure}

A finer decomposition resolves how the on-axis prediction is assembled from writes that individually point nowhere near it, and in particular how the residual reaches the read-out although the feed-forward writes that build it sit near $55^\circ$ from it (Section~\ref{sec:baseline-offaxis}). Attributing the final logits to each sublayer write, with the final normalization shared so that the writes sum to the logits exactly, the answer is written almost entirely by the last few feed-forward layers, the final one supplies about a third of the total, while attention and the token embedding contribute essentially nothing to the answer direction (Figure~\ref{fig:dla}a). The prediction emerges by coherent superposition (Figure~\ref{fig:dla}b): the answer-aligned parts of the writes share sign and accumulate, growing with depth, while the orthogonal parts, each write promoting its own other tokens, point in different directions and largely cancel, falling at the final write to $0.18$ times the random-walk level. The residual arrives on the read-out because the answer superposes and the rest cancels, not because any layer turns onto it.

\subsection{Why the frame rotates}
\label{sec:baseline-rotation}

Section~\ref{sec:baseline-additive} established that the off-axis constellation moves as a near-rigid body from layer to layer while the answer arrives on-axis by addition. That motion is worth characterizing on its own, because what is moving is the frame the network computes in. Nothing pins that frame in the baseline, and every layer lands on one of its own; only when a frame is supplied (Section~\ref{sec:device}) does the network settle into a fixed one for each phase. By a \emph{frame} we mean nothing more than a choice of coordinate directions for the off-axis subspace: which direction counts as the first axis, which as the second, and so on. The network's function does not depend on that choice. A measurement of what a given direction means does, which is why a frame that moves every layer matters. From one layer to the next the whole concept constellation turns by about $29^\circ$, carried by a single shared rotation, and the motion is rigid and input-independent: a rotation fitted only to the token stream and never to the concepts carries the constellation from layer to layer at $R^2 = 0.93$ against $0.76$ for leaving it unchanged (Section~\ref{sec:baseline-additive}), and one global rotation per transition retains $98$--$99\%$ of that agreement across disjoint inputs. The turn stays within the off-axis subspace, so the angle to the read-out holds flat through the concept phase while the frame beneath it rotates.

The turn is a property of the layer's linear response. Write the best linear description of a layer's write as $w \approx A\,h$; the frame carried across the transition is the orthogonal factor of $I + A$. A symmetric $A$ would stretch the constellation without turning it, so the turn measures how far $A$ departs from symmetry, and that departure is close to what an arbitrary matrix gives: the antisymmetric share of $A$'s energy is $0.675$--$0.687$, against $0.7064$ for a random matrix. The same operator has been studied spectrally, where a full Jacobian eigendecomposition shows early layers non-normal and rotation-dominated and late layers approaching symmetry \citep{fernando2026dynamics}; the polar factor used here isolates the angle that decomposition leaves implicit.

Two properties of a transformer layer make $A$ asymmetric. The read and write projections are untied, so their product has no reason to be symmetric. And the layer is nonlinear while the distribution of residual states it acts on is not Gaussian, which makes the best linear description of the layer differ from its average local slope in a way that carries no symmetry.\footnote{The distinction is between two different linear objects. The fit above is the best linear \emph{predictor} of the write, $A = \mathbb{E}[w h^{\top}]\,\Sigma^{-1}$ with $\Sigma = \mathrm{Cov}(h)$, while the layer's average local slope is the mean Jacobian $\mathbb{E}[\partial w / \partial h]$. Stein's lemma makes the two coincide exactly when the input is Gaussian,
\[
h \sim \mathcal{N}(\mu, \Sigma) \;\Longrightarrow\; \mathbb{E}\!\left[w h^{\top}\right]\Sigma^{-1} \;=\; \mathbb{E}\!\left[\partial w / \partial h\right],
\]
and the residual stream is not Gaussian, so they come apart: measured on the baselines the two agree at $\cos = 0.44$ on real states and at $0.98$ on synthetic states matched in mean and covariance. That least squares recovers the true direction only under an elliptical design, and departs from it by a bounded amount otherwise, is classical \citep{liduan1989,duanli1991}; the Gaussian-surrogate linearization of a transformer feed-forward block has been derived in closed form on the same identity \citep{belrose2025polynomials}.} Its effect on the frame is a matter of direction rather than size: matching the states to a Gaussian turns the antisymmetric part of $A$ by about $65^\circ$ while shrinking it by less than a fifth, so the nonlinearity sets where the frame turns more than how far.

The two sublayers contribute on different schedules. Measuring each sublayer's own contribution, attention's is flat across depth while the feed-forward path's grows steadily, from roughly a fifth of attention's at the first transition to several times it at the last. The two are not additive and the per-sublayer summary is not on the same scale as the per-layer figure above, so we report the shape rather than a decomposition: attention is the steady contributor and the feed-forward path is the one that grows with depth, which places the increase where the commit is.

\begin{table}[H]
\centering
\begin{tabular}{lc}
\toprule
property of the layer-to-layer turn & baseline value \\
\midrule
turn per layer & $29^\circ$ \\
rigidity, rotation fitted to the token stream alone & $R^2 = 0.93$ \;(vs $0.76$ unchanged) \\
retained by one global rotation per transition & $98$--$99\%$ \\
antisymmetric share of $A$ & $0.675$--$0.687$ \;(random matrix $0.7064$) \\
best linear response vs.\ mean Jacobian, real states & $\cos = 0.44$ \\
same, states matched in mean and covariance & $\cos = 0.98$ \\
antisymmetric part reoriented under Gaussian matching & ${\approx}\,65^\circ$ \\
its magnitude under Gaussian matching & $0.82\times$ \\
\midrule
\multicolumn{2}{l}{\emph{per-sublayer contribution (a different summary; not additive with the above)}} \\
attention, across depth & flat \\
feed-forward, across depth & grows ${\approx}\,4\times$ from first transition to last \\
\bottomrule
\end{tabular}
\caption{The layer-to-layer turn, measured on the converged baselines. The frame the concept phase computes in rotates rigidly with depth, and that rotation is the orthogonal factor of the layer's linear response. A layer whose response were symmetric would leave the frame fixed, and the response departs from symmetry by about as much as an arbitrary matrix does. The first block of rows is the per-layer turn of the concept constellation; the last two are a per-sublayer summary on a different scale, given for shape rather than as a decomposition.}
\label{tab:rotation}
\end{table}

The turn is small on the scale that governs this measurement, which is worth checking because angles between high-dimensional objects concentrate. A rigid $29^\circ$ rotation in $d = 768$ gives $\lVert R - I\rVert_F = \sqrt{2d(1-\cos 29^\circ)} \approx 13.9$, nearly three times below the $\sqrt{2d} \approx 39$ that two unrelated orthogonal matrices produce, so what we measure sits well inside the concentration scale rather than at it. Procrustes decompositions of transformer depth have been reported for token clouds \citep{bhattacharya2026residual}; the quantity here is the motion of the concept directions themselves, measured against an explicit no-motion null.

This motion leaves the rest of the picture untouched. The turn moves the frame within the off-axis subspace, so the angle to the read-out, the participation ratio, and the additive character of the commit are all unchanged by it.

\subsection{A usable baseline is roughly a coin flip}
\label{sec:baseline-convergence}

\begin{figure}[H]
\centering
\includegraphics[width=0.98\textwidth]{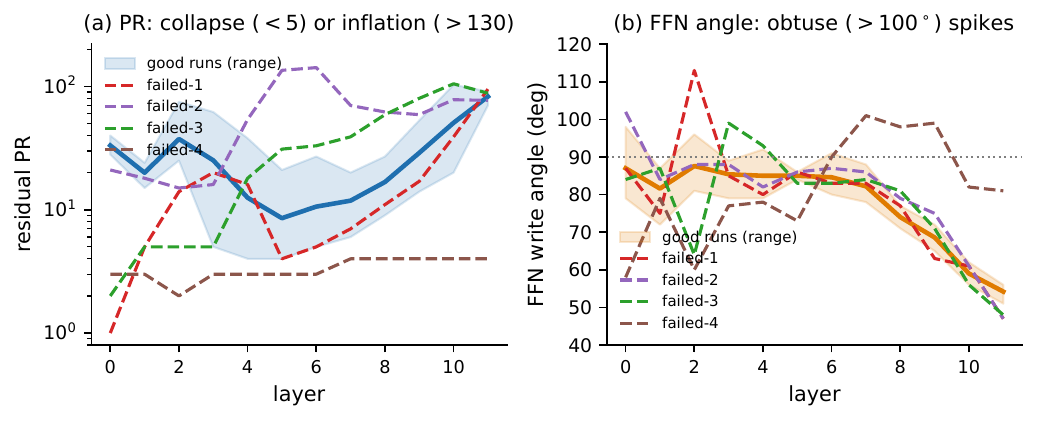}
\caption{The geometry of a usable model, and of the ones that fail (five converged, four unusable). (a)~Residual participation ratio: the converged runs (band) hold a moderate-then-expanding profile, while the four failed runs collapse below five, inflate past one hundred thirty, or sit flat near rank three. (b)~Feed-forward write angle: the converged band commits smoothly, while the failures carry obtuse ($>\!100^\circ$) writes that point away from the read-out.}
\label{fig:convergence-geom}
\end{figure}

The seed sweep also answers the reliability question, and ties it to the geometry. Five of nine seeds produced a usable model. The unusable runs share no single failure mode. Figure~\ref{fig:convergence-geom} sets the converged envelope against all four: one collapses to near rank-one in its early layers (participation ratio one to five), another inflates through the middle of the network (above one hundred thirty), and write angles turn obtuse, past $100^\circ$, pointing away from the read-out, in each failure and in none of the converged runs. The late feed-forward swing toward the read-out survives in the failures, so a missing commit is not what separates them; the signature of a usable model is the intact dimensionality profile, moderate through the body, expanding at the output, together with the absence of anti-read-out writes. Convergence is the event of landing in this geometry, and at this scale and training budget it happens about half the time.

\begin{table}[H]
\centering
\begin{tabular}{lcccc}
\toprule
outcome & seeds & LAMBADA & BLiMP & validation perplexity \\
\midrule
converged & $5/9$ & $0.266$ \;\; [$0.254$--$0.271$] & $0.805$ \;\; [$0.800$--$0.808$] & $19.17$ \;\; [$19.00$--$19.65$] \\
unusable & $4/9$ & $0.00$--$0.21$ & $0.53$--$0.80$ & $21.6$--$39.5$ \\
\bottomrule
\end{tabular}
\caption{Outcomes across nine seeds trained with identical hyperparameters. Converged runs are reported as mean~[range]; benchmark levels for models of this class are consistent with prior reports \citep{oskin2026legible}.}
\label{tab:baseline-runs}
\end{table}

\section{Forcing GELU on-axis}
\label{sec:forcing}

Section~\ref{sec:baseline} showed that a trained GELU transformer computes off the read-out by default. Whether that off-axis organization is necessary, or merely what the network settles into, is a question we can answer directly: pressure every layer onto the read-out and measure what the model does, how close it gets, and what it costs.

\subsection{Pressure toward the read-out}
\label{sec:forcing-method}

To pull the computation on-axis we add a training penalty that rewards every layer for decoding to the answer. Write $z^{(\ell)}_t = \mathrm{LN}(h^{(\ell)}_t)\,E^{\top}$ for the logit-lens logits of layer $\ell$'s residual at position $t$: the residual sent through the final layer norm $\mathrm{LN}$ and the tied token embedding $E$, as in Section~\ref{sec:methods}, and $z_t$ for the model's final logits. The penalty at coefficient $\alpha$ is the mean over the $L$ layers of the divergence between each layer's decoded distribution and the final one,
\begin{equation}
\mathcal{L}_{\mathrm{KL}} \;=\; \frac{\alpha}{L}\sum_{\ell=1}^{L}\mathbb{E}_{t}\Big[\,D_{\mathrm{KL}}\!\big(\,\mathrm{softmax}(z_t)\ \big\|\ \mathrm{softmax}(z^{(\ell)}_t)\,\big)\,\Big],
\label{eq:kl}
\end{equation}
with the target $\mathrm{softmax}(z_t)$ held fixed (stop-gradient), the expectation over a subsample of token positions, and $\mathcal{L}_{\mathrm{KL}}$ added to the cross-entropy during training and switched off at evaluation. Minimizing it makes each layer decode to the final prediction, which is to say it places each layer on the read-out. The coefficient $\alpha$ sets how hard we push; we sweep it from light ($0.1$) to heavy ($10$) and, following Section~\ref{sec:methods}, report the converged runs.

\begin{figure}[H]
\centering
\includegraphics[width=0.92\textwidth]{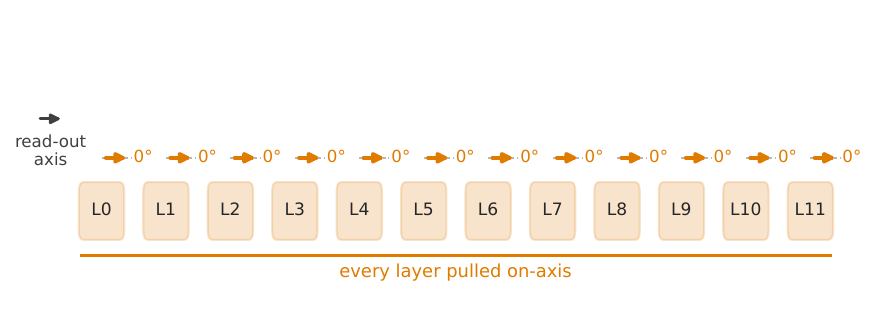}
\caption{The pressure applied in this section. Every layer's decode is asked to point along the final prediction: a target angle of $0^\circ$ at all twelve depths, with nothing inserted into the residual stream. Each arrow is the \emph{prescribed} target for that layer, measured from the read-out direction; what the network actually does under this pressure is Figure~\ref{fig:forced-frontier}.}
\label{fig:sched-forcing}
\end{figure}

\subsection{Closer to the axis, at a rising cost}
\label{sec:forcing-cost}

\begin{figure}[H]
\centering
\includegraphics[width=0.98\textwidth]{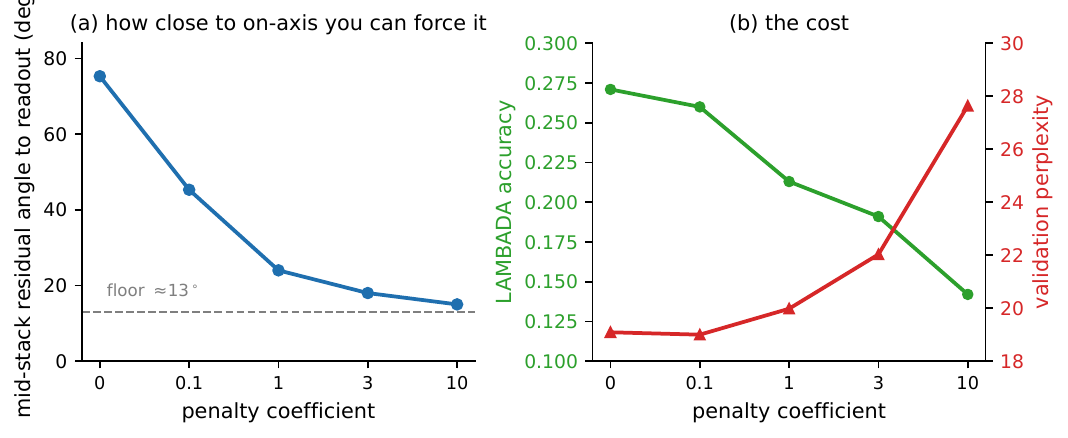}
\caption{Forcing the computation on-axis with the read-out-alignment penalty, swept in coefficient. (a)~The mid-stack residual angle to the read-out falls with pressure and saturates toward a $\approx\!13^\circ$ floor. (b)~LAMBADA accuracy falls and validation perplexity climbs across the same sweep. Closer to the axis is monotonically worse; the baseline is the best point.}
\label{fig:forced-frontier}
\end{figure}

The penalty moves the computation on-axis, and it saturates (Figure~\ref{fig:forced-frontier}a). The mid-stack residual, $75^\circ$ off the read-out in the baseline, is drawn to $45^\circ$, then $24^\circ$, $18^\circ$, and $15^\circ$ as the coefficient climbs, approaching but never reaching a floor near $13^\circ$. Quality falls the whole way (Figure~\ref{fig:forced-frontier}b): LAMBADA drops from $0.271$ to $0.142$ and perplexity climbs from $19$ to $28$. The two move together, each step toward the read-out costs more of the model, so there is no coefficient at which being closer to the axis yields a better result. The baseline, farthest off-axis, is the best point on the curve. Pressing GELU toward the read-out buys proximity and pays in quality.

\subsection{The answer survives as a residue of cancellation}
\label{sec:forcing-contortion}

The penalty pays in quality because it asks the model to match the answer at every layer and the model cannot, so it degrades. A blunter pressure evades the trap. Rather than match distributions, constrain the geometry directly: penalize the \emph{angle} of each layer's decode. With the logit vectors centered over the vocabulary, $\tilde{z} = z - \bar{z}\,\mathbf{1}$, and a target angle $\theta_\ell$ per layer, the penalty is
\begin{equation}
\mathcal{L}_{\cos} \;=\; \frac{\alpha}{L}\sum_{\ell=1}^{L}\mathbb{E}_{t}\Big[\big(\cos\angle(\tilde{z}^{(\ell)}_t,\ \tilde{z}_t) - \cos\theta_\ell\big)^2\Big],
\qquad \cos\angle(a,b)=\frac{\langle a,b\rangle}{\lVert a\rVert\,\lVert b\rVert},
\label{eq:cos}
\end{equation}
where forcing on-axis sets $\theta_\ell = 0$ at every layer, so the target is $\cos 0 = 1$ and each layer's centered decode is driven to point along the final prediction. Under this loss the residual reaches the same $\approx\!13^\circ$ floor and the model trains to near-baseline quality. This is not a single-seed accident: we trained fifteen seeds under the angle constraint, eight reached the end of training, and the six that produced usable models average LAMBADA $0.258$ and perplexity $19.5$, against the distributional penalty's $0.191$ and $22.0$ at the same coefficient. On the numbers, and across seeds, forcing GELU on-axis looks free.

\begin{figure}[H]
\centering
\includegraphics[width=0.98\textwidth]{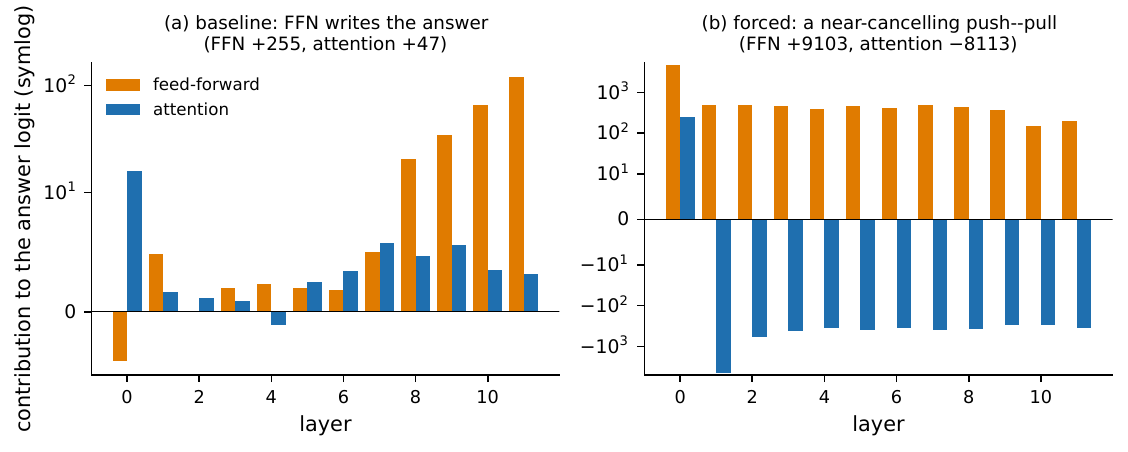}
\caption{How the on-axis-forced model reaches the answer, by direct logit attribution (symmetric-log scale; note the different vertical scales). This is the hard $0^\circ$ angle target at coefficient $3$, averaged over its six converged seeds (near-baseline quality, LAMBADA $0.258$ / perplexity $19.5$). (a)~In the baseline the feed-forward path writes the answer, late, and attention contributes almost nothing. (b)~Forced on-axis, both sublayers write enormous, nearly-cancelling contributions at every layer, feed-forward $+9103$ toward the answer, attention $-8113$ against it, so the answer survives only as the residue of a near-total cancellation.}
\label{fig:forced-contortion}
\end{figure}

It is not free. Attributing the final prediction to each write shows what the model built to satisfy the constraint (Figure~\ref{fig:forced-contortion}). Averaged over the six seeds the feed-forward path over-writes the answer with a contribution of $+9103$ to the answer logit, and attention writes an almost-equal correction against it, $-8113$; the two nearly cancel, leaving the small surviving answer. The individual magnitudes vary by seed, but the cancellation does not: attention supplies about $-0.9$ times the feed-forward answer-mass and is net-negative in every one of the six runs. The coherence of the answer-writing collapses from $0.71$ in the baseline to $0.07$, and the opposing contributions inflate by a factor of thirty-six, the feed-forward total rising from $+255$ in the baseline to $+9103$, with the final normalization rescuing a sane perplexity from the near-cancellation. The sublayers split geometrically to match: the feed-forward write locks on-axis at $\approx\!14^\circ$ while attention flips anti-aligned to $\approx\!150^\circ$.

This is the blur of Section~\ref{sec:baseline-offaxis} made visible. Attention mixes across token positions, so on the read-out it would smear the prediction; the only way to hold the answer sharp while forcing attention on-axis is to make it write \emph{against} the answer, cancelling the blur it would otherwise inflict. The contortion is the baseline's insulation defending itself under duress: the network keeps the read-out clear of cross-token mixing even when every layer is pressed onto it. The workspace itself is not collapsed: the residual participation ratio is reduced but stays well clear of rank collapse, which appears only in runs that fail to train, so this is not a broken model but a working one that reaches the answer by a grotesque cancellation rather than by computing it on-axis. Forcing GELU onto the read-out is possible; the model it produces is a caution, not a solution.

Being on-axis is thus achievable, but not with a model that behaves normally. A GELU transformer needs to compute off the read-out, where attention can mix across positions without blurring the answer. In the next section we stop forcing every layer to one axis and force a phase structure instead: an off-axis phase and then an on-axis one, exploring which axis to demand, and where.

\section{Prescribing the phase structure}
\label{sec:step}

Section~\ref{sec:forcing} forced every layer onto the read-out and found it possible but grotesque: the model reaches the answer by a near-total cancellation rather than by computing it on-axis, because attention cannot mix on the read-out without blurring the prediction. The baseline avoids this by splitting its depth into two phases: an off-axis concept phase where attention mixes freely, and a late on-axis commit. That split is the organization we want to reproduce; the question this section asks is whether we can \emph{prescribe} it. The baseline discovers the phase structure on its own, easing onto the axis over its final layers; here we demand it sharply, as a step, and through the training loss alone.

\subsection{The step schedule}
\label{sec:step-method}

We reuse the angle penalty of Equation~\ref{eq:cos}, but with a per-layer target that encodes the two phases rather than a single on-axis goal. In words, the loss asks the first half of the network to keep everything it computes invisible to the unembedding, and the second half to point at the answer. The first half of the network is pinned off-axis and the second half on-axis,
\begin{equation}
\theta_\ell = \begin{cases} 90^\circ, & \ell = 0,\dots,5 \quad\text{(concept phase)}\\[2pt] 0^\circ, & \ell = 6,\dots,11 \quad\text{(token phase)}, \end{cases}
\label{eq:step}
\end{equation}
so the loss rewards each concept-phase layer for keeping its decode orthogonal to the answer and each token-phase layer for pointing at it. Nothing else changes: no architectural modification and no inserted rotation, the phase structure is requested through the loss and must be built by the weights. We call this the \emph{frameless} step, to set it apart from the rotation device introduced later, which hands the network an off-axis frame directly rather than asking it to grow one. We train eight seeds and, following Section~\ref{sec:methods}, report the converged set.

\begin{figure}[H]
\centering
\includegraphics[width=0.92\textwidth]{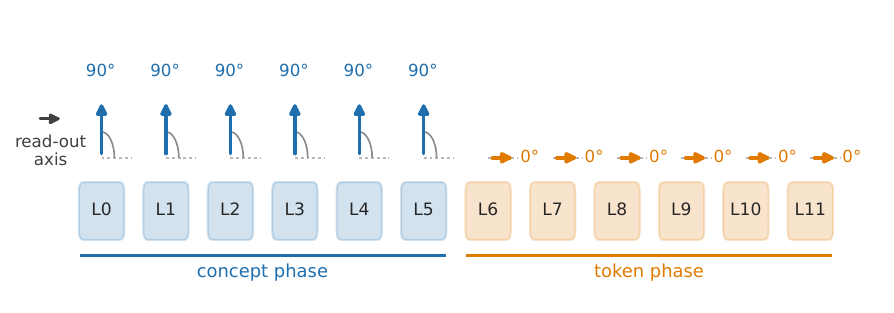}
\caption{The pressure applied in this section. The same angle penalty as Section~\ref{sec:forcing}, but with a per-layer target that encodes two phases rather than one on-axis goal: $90^\circ$ through $L0$--$L5$ and $0^\circ$ through $L6$--$L11$ (Equation~\ref{eq:step}). Nothing is inserted into the residual stream, so the boundary is requested through the loss and has to be built by the weights. Arrows are prescribed targets, not realized geometry.}
\label{fig:sched-step}
\end{figure}

\subsection{A lottery}
\label{sec:step-lottery}

The frameless step is unreliable in a way the uniform on-axis force was not. Of eight seeds, six diverged outright, aborting at instantaneous perplexities between roughly $110$ and $3000$; there is no middle ground between the failures in the hundreds and the completions near perplexity $20$--$33$. The instability is not an early-training artifact: one seed trained stably at perplexity $\approx\!33$ to eighty percent of the schedule before a sudden late spike, and a warm restart rescued one of the six failures while failing to rescue another. Prescribing the phase structure by loss alone reaches the end of training about a quarter of the time.

\begin{figure}[H]
\centering
\includegraphics[width=0.98\textwidth]{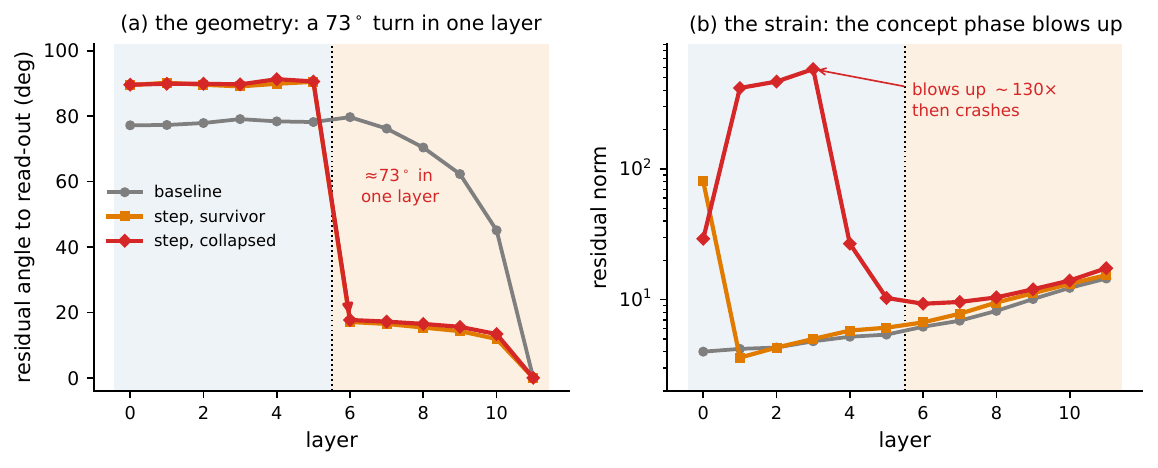}
\caption{The frameless step realizes the prescribed geometry but strains before the boundary. (a)~Residual angle to the read-out across depth: the baseline eases onto the axis over its final layers, while both completed step seeds hold $\approx\!90^\circ$ through the concept phase and turn $\approx\!73^\circ$ onto the read-out in the single layer at the split. The geometry does not separate the survivor from the collapsed run. (b)~Residual norm (log scale): the survivor absorbs one first-layer blast and then tracks the baseline, while the collapsed run blows up $\sim\!114\times$ through layers one to three and crashes back before the boundary. The strain lives in the norm, not the angle.}
\label{fig:step-violence}
\end{figure}

The two seeds that trained to completion do not agree on the result. One reached baseline quality: the \emph{survivor}, LAMBADA $0.237$, perplexity $20.6$; the other limped to the boundary of usability, the \emph{collapsed} run, LAMBADA $0.161$, perplexity $33.0$. Yet on the prescribed geometry they are indistinguishable. Figure~\ref{fig:step-violence}a shows the residual's angle to the read-out across depth: where the baseline eases toward the axis over its last several layers, both step models hold $\approx\!90^\circ$ flat through the concept phase and then snap $\approx\!73^\circ$ onto the read-out in the single layer at the boundary. The target angle is met either way; the geometry does not tell the two apart.

What separates them is the strain of meeting it, and the strain is spent before the boundary, not at it. Figure~\ref{fig:step-violence}b tracks the residual norm. The surviving seed absorbs a single blast at the first layer and then tracks the baseline; the collapsed seed's residual norm balloons to $114\times$ the baseline's, peaking near $580$, through layers one to three, then crashes back toward the baseline before it ever reaches the boundary. The violence of the frameless step is not the on-axis snap itself, which every write at the boundary is small enough to perform. It is the pre-boundary contortion the concept phase undergoes in order to arrive at the boundary still held off-axis.

\begin{figure}[H]
\centering
\includegraphics[width=0.66\textwidth]{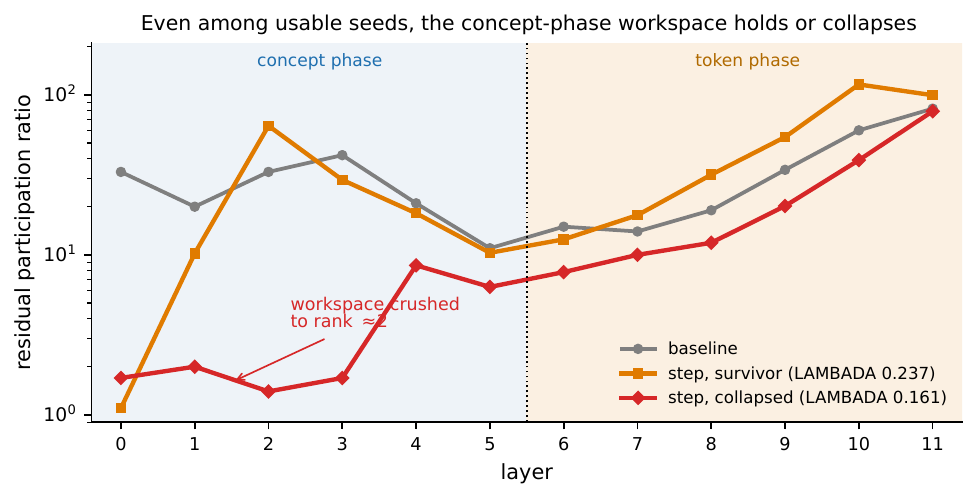}
\caption{Among the seeds that converge, the concept-phase workspace either holds or collapses. Residual participation ratio across depth (log scale) for the baseline and the two completed step seeds. The survivor (orange) expands its workspace through the concept phase as the baseline does; the collapsed run (red) is crushed to rank~$\approx\!2$ through the first four layers: the phase the step pins off-axis, and recovers only after the commit. The loss prescribes the angle, not the room to compute at it.}
\label{fig:step-lottery}
\end{figure}

That contortion is where the lottery is decided, and it is decided inside the concept phase. Figure~\ref{fig:step-lottery} gives the residual participation ratio across depth for the same three models. The surviving seed carries a healthy workspace, low at the first layer but expanding immediately, much as the baseline's does, while the collapsed seed's workspace stays crushed near rank~$2$ through layers zero to three, exactly the layers the step pins off-axis, recovering only in the token phase after the commit. Holding the concept phase at $90^\circ$ by loss alone either coexists with a full-rank workspace or crushes it, and the angle loss cannot tell the model which: the target says where to point but nothing about keeping room to compute while pointing there. The collapsed run pays for losing that room exactly as Section~\ref{sec:baseline-offaxis} predicts, the off-axis workspace it forfeits is the functional insulation whose removal was $64$--$84\times$ more costly than a random rotation, and its quality falls to the boundary of usability.

The frameless step therefore reproduces the baseline's phase structure only by luck. The geometry is cheap to name and cheap to hit, both completions sit on it, but the off-axis workspace that makes the geometry \emph{useful} is not something the angle loss can supply; the model must build it under strain, and most of the time the strain wins. What the loss prescribes is a direction; what a working concept phase needs is a frame: a stable off-axis subspace with room to mix. The next section supplies that frame directly, as a fixed rotation inserted into the residual stream, and asks whether the prescription that was a lottery becomes reliable once the model is handed the frame instead of made to grow one.

\section{Supplying the frame}
\label{sec:device}

Section~\ref{sec:step} left the frameless step as a lottery: prescribing the phase structure through the loss asks the model to grow an off-axis workspace under strain, and most of the time the strain wins. The diagnosis was specific: the loss names a direction but supplies no frame, so the model must both discover a stable off-axis subspace and hold it against the pressure. This section removes that burden. Rather than ask the network to build the off-axis frame, we hand it one: a fixed rotation inserted into the residual stream at the concept/token boundary, so the concept phase can compute in a frame the device then turns onto the read-out.

\subsection{The device}
\label{sec:device-method}

The device is a single fixed orthogonal rotation $R$ applied to the residual stream once, at the boundary $\ell = 6$:
\begin{equation}
h^{(6)} \;\leftarrow\; R\,h^{(6)}, \qquad R = Q^{\top} R_{90}\, Q,
\label{eq:device}
\end{equation}
where $R_{90}$ is a block-diagonal quarter-turn ($R_{90}^2 = -I$) and $Q$ is a fixed Haar-random orthogonal matrix. Conjugating by a dense $Q$ spreads the rotation across all $d$ coordinates, giving the \emph{dense} device; the sparser forms of the next section restrict where $R$ acts. The rotation is not learned and does not depend on the input, and because it is orthogonal it preserves the residual norm and the participation ratio exactly: it reorients the frame without touching the workspace. We train with the step schedule of Equation~\ref{eq:step} still in place, so the target geometry is unchanged from Section~\ref{sec:step}; the only difference is that the network now inherits the boundary rotation instead of having to manufacture it. The picture the device makes concrete is the one the baseline already follows and Section~\ref{sec:forcing} confirmed under duress: the model computes off the read-out and turns onto it, never the reverse. The device performs that turn, at the boundary, for free.

\begin{figure}[H]
\centering
\includegraphics[width=0.92\textwidth]{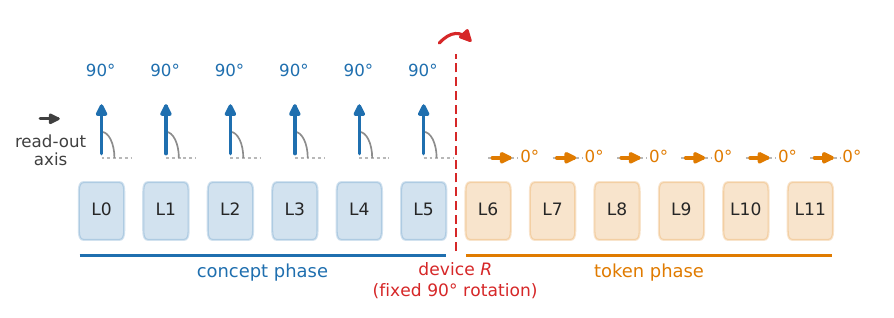}
\caption{The pressure applied in this section, and the device. The schedule is unchanged from Figure~\ref{fig:sched-step}: $90^\circ$ through $L0$--$L5$, $0^\circ$ through $L6$--$L11$. The single addition is the fixed rotation $R$ applied to the residual stream once, at the boundary $\ell = 6$ (Equation~\ref{eq:device}), so the network inherits the boundary turn instead of having to manufacture it. Arrows are prescribed targets, not realized geometry.}
\label{fig:sched-device}
\end{figure}

It may help to state what this does in ordinary terms before the measurements. The concept phase needs somewhere to compute that the unembedding ignores, and Section~\ref{sec:step} showed that it will not reliably find such a place on its own. The device hands it one. Everything before the boundary is written in a frame that the device is going to rotate, so the concept phase has a fixed target to compute against rather than a target it must invent and hold. At the boundary the rotation turns that frame onto the read-out, and the layers after it read the result and adjust to the new orientation, which they can do because they are trained with the rotation already in place.

That turn puts concept content on the read-out axis, which is the position Section~\ref{sec:baseline-offaxis} showed to be expensive, and it is worth saying why the two situations differ. The insulation result is about \emph{averaging}: the damage came entirely from letting attention mix values that already carried vocabulary predictions, and the matched control, the same rotation with the cross-token averaging removed, cost about as much as a random rotation of the same size. The device performs no averaging. It applies one fixed orthogonal map to each position independently, which is precisely the harmless arm of that control. What arrives on the read-out after the boundary is concept content rather than a set of competing next-token predictions, and the token phase writes the answer on top of it additively, as Section~\ref{sec:baseline-additive} showed the untouched network already does.

\subsection{The lottery is gone}
\label{sec:device-lottery}

With the frame supplied, the prescription lands. Seven of eight dense-device seeds trained to a usable model, against two of eight for the frameless step, and the survivors are not marginal: they sit at baseline quality, LAMBADA $0.257$, perplexity $19.3$, BLiMP $0.798$, all within the converged baseline's range (Table~\ref{tab:device}). The step schedule that was a coin flip in Section~\ref{sec:step} becomes routine once the model is not also asked to grow the frame it is being held to. The device buys nothing in raw quality, nor could it, since the baseline already reaches that quality on its own, but it converts an unreliable prescription into a reliable one, and that reliability is the result.

\begin{table}[H]
\centering
\begin{tabular}{lccccc}
\toprule
regime & converged & LAMBADA & BLiMP & perplexity & concept-phase PR \\
\midrule
baseline (\S\ref{sec:baseline}) & $5/9$ & $0.266$ \;[$.254$--$.271$] & $0.805$ & $19.2$ & $24$ \\
forced on-axis (\S\ref{sec:forcing}) & $6/15$ & $0.258$ & ${\approx}0.80$ & $19.5$ & $14$ \;[$9$--$21$] \\
frameless step (\S\ref{sec:step}) & $2/8$ & $0.161$--$0.237$ & $.69$--$.79$ & $20.6$--$33.0$ & $4$--$22$ \\
\textbf{dense device} (\S\ref{sec:device}) & $\mathbf{7/8}$ & $\mathbf{0.257}$ \;[$.226$--$.269$] & $\mathbf{0.798}$ & $\mathbf{19.3}$ & $\mathbf{54}$ \;[$42$--$64$] \\
\bottomrule
\end{tabular}
\caption{The device removes the lottery. Convergence (usable models / seeds), quality, and the mean concept-phase participation ratio (over layers~0--6) across the four regimes of Sections~\ref{sec:baseline}--\ref{sec:device}. Quality is flat across every non-collapsed regime; what changes is \emph{reliability} and the \emph{workspace}. Forcing on-axis (\S\ref{sec:forcing}) survives by contortion at a reduced-but-intact rank; the frameless step (\S\ref{sec:step}) is a lottery between collapse and survival; only the device delivers reliable convergence together with a full-dimensional off-axis workspace, at baseline quality. Converged runs reported as mean~[range]; PR mean over the converged set.}
\label{tab:device}
\end{table}

\subsection{The geometry, with room to compute}
\label{sec:device-geometry}

\begin{figure}[H]
\centering
\includegraphics[width=0.98\textwidth]{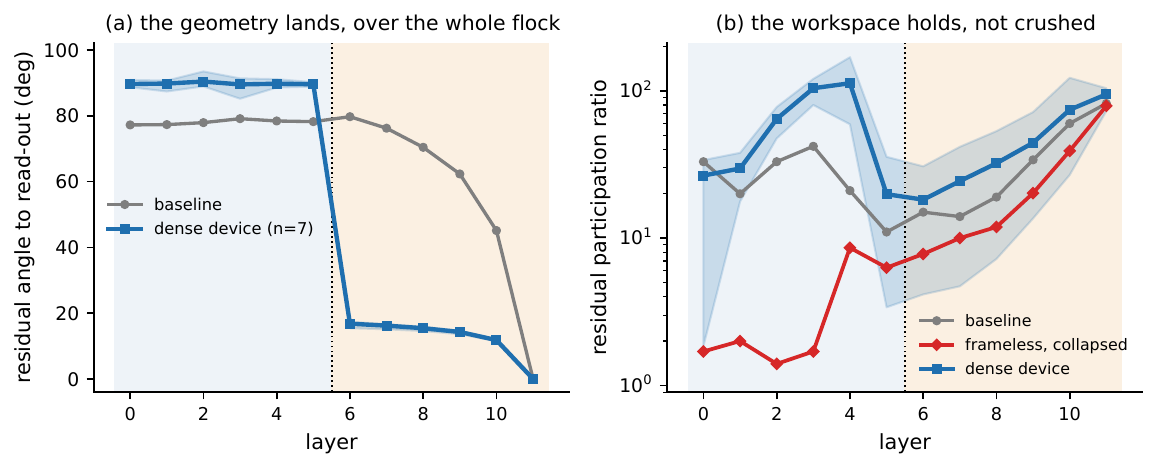}
\caption{The device lands the geometry and keeps the room (converged dense-device flock; band spans the flock, bold the mean). (a)~Residual angle to the read-out across depth: the concept phase holds $\approx\!90^\circ$ and turns onto the read-out at the boundary, over the whole converged set rather than the lucky seeds of Section~\ref{sec:step}. (b)~Residual participation ratio: the concept-phase workspace stays richly dimensional (mean ${\approx}54$, peaking above $100$), against the frameless step's collapse to rank $\approx\!2$ (red) and the baseline (grey). The device supplies the full-dimensional off-axis frame the loss alone could not. The frame in which this computation happens is itself turning with depth: Section~\ref{sec:baseline-rotation} shows that turn is the orthogonal part of each layer's linear response, that it is largest where concepts are being formed rather than where the answer is written, and that it follows from a nonlinear map acting on a residual distribution that is not Gaussian.}
\label{fig:device}
\end{figure}

The device reproduces the prescribed geometry, and, unlike the frameless step, it does so with the workspace intact (Figure~\ref{fig:device}). Across the converged flock the residual holds $\approx\!90^\circ$ off the read-out through the concept phase and turns onto it at the boundary (Figure~\ref{fig:device}a), the same two-phase profile the survivor and the collapsed seed of Section~\ref{sec:step} both realized, but here it is the behavior of the whole converged set, not the lucky half. The difference is in the participation ratio (Figure~\ref{fig:device}b): the concept-phase workspace stays richly dimensional, a mean participation ratio of about $54$ across the flock, peaking above $100$ mid-phase, roughly twice the baseline's, where the frameless step's collapsed seed was crushed to rank $\approx\!2$ through exactly those layers. The device supplies the full-dimensional off-axis frame the frameless step could not grow, and the concept phase computes in it.

This is the payload of the section, and it is what the exact-$90^\circ$ concept phase means. Holding the concept phase orthogonal to the read-out places its writes in a subspace the unembedding discards: the off-axis computation contributes nothing on-axis, so it adds no noise to the forming prediction. That is the insulation of Section~\ref{sec:baseline-offaxis} made a design, the same off-axis positioning whose violation cost $64$--$84\times$ a random rotation, now installed deliberately and held full-rank by the device rather than grown under strain. The device lets the model form concepts off the read-out with no on-axis leakage, at full dimensionality, reliably.

\subsection{The device supplies the frame}
\label{sec:device-control}

Section~\ref{sec:baseline-additive} established that rotation is not something the untouched network does at any depth or in either direction, which is what makes inserting one a genuine intervention rather than a nudge along a path the model already takes. One caution is worth ruling out in the other direction: that the device simply rotates the answer into place, doing the model's committing for it. It does not. The rotation the device supplies is, in isolation, a pure quarter-turn, fit it with the orthogonal Procrustes decomposition of Section~\ref{sec:baseline-additive} and it is recovered exactly ($R^2_{\mathrm{rot}} = 1.00$). Yet the residual transition \emph{across the boundary}, device and all, is no more a rotation than the baseline's own commit: a best-fit rotation captures the same minority fraction ($R^2_{\mathrm{rot}} \approx 0.38$, against the baseline's $0.36$--$0.39$), the rest freshly written content. The network writes its answer additively \emph{through} the device's frame, late and on-axis, exactly as the baseline does; the device contributes the frame, not the commit. Nor is the realized geometry an artifact of decoding: applying the same rotation $R$ post-hoc to a trained baseline does not reproduce the device model's profile. It turns the angles the wrong way, \emph{away} from the read-out, throwing the final-layer residual from on the read-out ($0^\circ$) to $83^\circ$ off it. The trained network places its content on the prescribed geometry; the device only fixes the frame it is placed in. What the model computes and when it commits are unchanged: the device changes only the coordinate system the concept phase lives in, and by doing so makes the off-axis phase something the network can be handed rather than forced to grow.

\begin{figure}[H]
\centering
\includegraphics[width=0.78\textwidth]{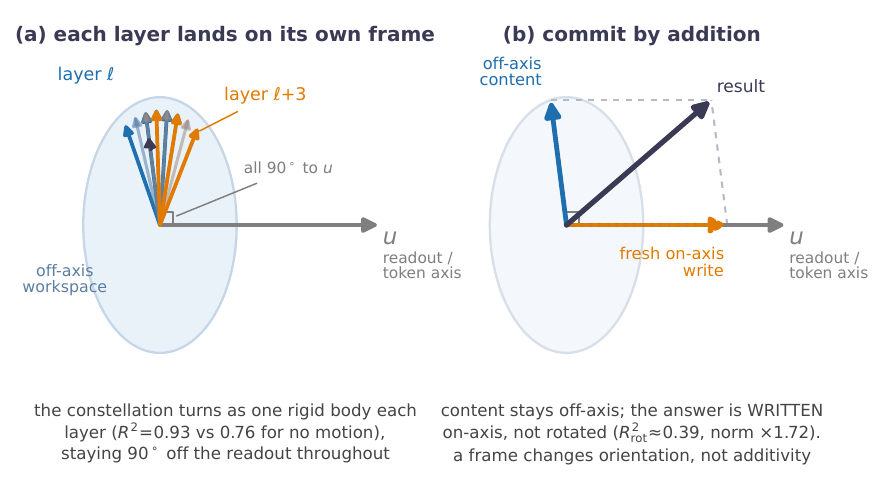}
\caption{Two motions, told apart, and both survive the device. (a)~In the baseline every layer lands on a frame of its own: one rigid rotation carries the whole constellation across each transition ($R^2 = 0.93$ for a rotation fitted only to the token stream, against $0.76$ for no motion), with every direction staying $\approx\!90^\circ$ off the read-out. The turn is drawn small; it happens in a plane normal to the read-out. (b)~The commit instead adds a fresh on-axis component to off-axis content carried unchanged, so the result leans on-axis by addition rather than rotation. Both behaviours are unchanged when the device supplies the frame.}
\label{fig:framemotions}
\end{figure}

Two things carry over from the baseline unchanged, and together they say what the device does and does not touch. The commit stays additive: across the boundary a best-fit rotation captures the same minority fraction it captures in the untouched network, so supplying a frame does not convert the answer into something the model turns onto the read-out (Section~\ref{sec:baseline-additive}). And the frame keeps moving. Prescribing the phase structure fixes two frames, one for the concept phase and one for the token phase, where the baseline had a different frame at every layer (Section~\ref{sec:baseline-rotation}); within the concept phase the constellation still turns from one layer to the next, by $28.7^\circ$ under the device against the baseline's $29.0^\circ$. The device sets which frame the concept phase computes in. It does not stop the residual turning inside it, because that turn is a property of each layer's own linear response, and inserting a fixed rotation at the boundary leaves those responses exactly as asymmetric as they were (Section~\ref{sec:baseline-rotation}).

The dense rotation used here is, deliberately, the most demanding form of the device: mixing every coordinate, it cannot be absorbed by the network's normalization and must be genuinely accommodated. The next section asks whether a cheaper frame, a sparse rotation the normalization can absorb, does the same job, and what, if anything, the richness of the frame is worth.

\section{A cheaper frame}
\label{sec:sparse}

The dense rotation of Section~\ref{sec:device} was chosen to be the hardest case: by mixing every coordinate it cannot be folded into the network's own parameters, so the model must accommodate it outright. That choice raises two questions, one practical and one conceptual. The practical: does a \emph{cheaper} frame, a rotation the network can absorb, do the same job? The conceptual: both rotations are full-rank and involve every coordinate, but one mixes them all while the other only pairs them, and that difference sets how widely the concept phase ends up spread. Does that spread matter?

\subsection{The sparse device}
\label{sec:sparse-method}

The sparse device rotates within adjacent coordinate pairs rather than across all of them. Writing the residual in consecutive pairs $(x_{2i}, x_{2i+1})$, each pair is turned a quarter-circle,
\begin{equation}
(x_{2i},\, x_{2i+1}) \;\longmapsto\; (-x_{2i+1},\, x_{2i}),
\label{eq:sparse}
\end{equation}
so $R$ is block-diagonal in $2\times 2$ blocks: a signed permutation of the coordinates. As in Section~\ref{sec:device} it is fixed, applied once at the boundary, and trained under the same step schedule. What differs is what the network can do with it. A signed permutation only reorders channels and flips their signs, operations the layer normalization's per-channel scale and the surrounding weights can undo by relabeling; the network can absorb the sparse device almost for free, so it costs little to train around. The dense rotation admits no such relabeling, mixing all $d$ coordinates, it breaks the per-channel structure the normalization relies on, and the model must genuinely co-adapt to it. The gap is large and measurable: fitting the best per-channel affine to undo each device on a trained baseline leaves a relative residual of $0.05$ for the sparse device against $0.34$ for the dense one, averaged over the concept phase. The remaining $0.05$ is the cost of the mean subtraction, which is what separates the normalization used here from RMSNorm, where a signed permutation is an exact gauge symmetry \citep{sweeney2026signedperm}; repeating the measurement under RMSNorm drops the sparse residual to $0.002$ and leaves the dense one at $0.33$. The two devices ask the network for the same geometry; only one of them can be shrugged off.

\subsection{The same value, in a smaller workspace}
\label{sec:sparse-parity}

\begin{figure}[H]
\centering
\includegraphics[width=0.62\textwidth]{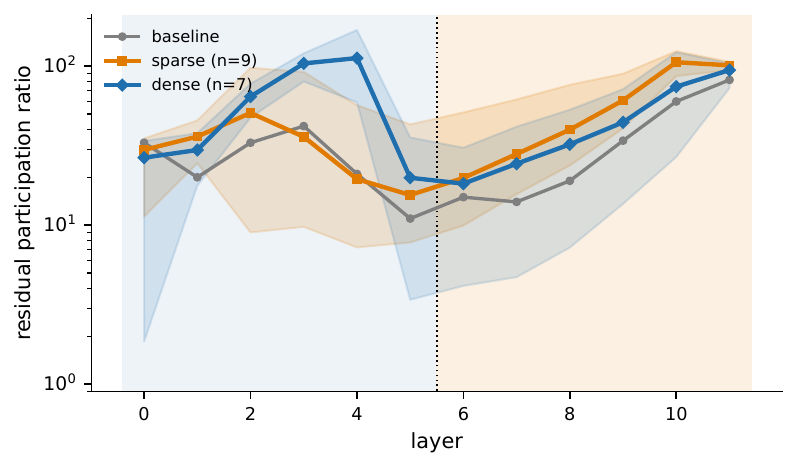}
\caption{The form of the frame changes the geometry, not the value. Concept-phase participation ratio across depth: the sparse device (orange) holds the workspace at about half the dimensionality of the dense device (blue), near the baseline (grey), while both realize the same $90^\circ$-then-commit angle profile. Quality is flat across all three (Table~\ref{tab:sparse}), and Figure~\ref{fig:gauge}a shows it run by run.}
\label{fig:sparse}
\end{figure}

The cheaper frame does the same job (Figure~\ref{fig:sparse}, Table~\ref{tab:sparse}). Sparse-device models reach the quality of the dense ones and of the baseline, LAMBADA $0.260$ against dense's $0.257$ and the baseline's $0.266$, with perplexity and BLiMP flat across all three (Table~\ref{tab:sparse}), and they realize the same two-phase geometry, holding the concept phase off the read-out and turning onto it at the boundary. Yet the workspace they compute in is markedly smaller: the concept-phase participation ratio is about $30$ for the sparse device, close to the baseline's own $24$, against about $54$ for the dense one (Figure~\ref{fig:sparse}), a roughly twofold difference in effective dimensionality that leaves quality untouched.

This is the section's point. The form of the frame sets the geometry, how many directions the concept phase is spread across, but not the value. A dense rotation scatters each concept over all $d$ coordinates and a sparse one keeps it within pairs, and the model is equally good either way. What the device must supply is a frame that clears the collapse of Section~\ref{sec:step}; how rich that frame is beyond the floor does not register in quality. Participation ratio, here, is not capability: the smaller workspace will do.

\begin{table}[H]
\centering
\setlength{\tabcolsep}{4pt}
\begin{tabular}{lccccc}
\toprule
device form & converged & LAMBADA & BLiMP & perplexity & concept-phase PR \\
\midrule
sparse (signed permutation) & $9/9$ & $0.260$ \;[$.245$--$.287$] & $0.794$ & $19.18$ & $30$ \;[$17$--$56$] \\
dense (Haar rotation) & $7/8$ & $0.257$ \;[$.226$--$.269$] & $0.798$ & $19.30$ & $54$ \;[$42$--$64$] \\
\bottomrule
\end{tabular}
\caption{Sparse versus dense device (converged runs; mean~[range], PR mean over layers~0--6). Quality is identical within variation; the workspaces the two devices induce differ twofold in effective dimensionality. The sparse device, which the network can absorb, also converges more reliably (see text). Baseline for reference: LAMBADA $0.266$, BLiMP $0.805$, perplexity $19.17$, concept-phase PR $24$.}
\label{tab:sparse}
\end{table}

\subsection{The consistency edge}
\label{sec:sparse-consistency}

If richness is free, the choice between the two forms comes down to how easily each trains, and here the sparse device has the advantage its absorbability predicts. Because a signed permutation can be relabeled away, it perturbs training little, and every sparse seed we ran reached a usable model. It reached it smoothly: no sparse run showed the sudden, unusually high spikes in instantaneous perplexity that the dense runs carry through training. The dense device, which the network must accommodate rather than absorb, produces those spikes, and one of its seeds diverged late. The difference is therefore visible in the training curves and not only in the final tally, which is what makes it a statement about how the two frames train. An absorbable frame is a gentler thing to train against than one that must be co-adapted, and that recommends the sparse device as the practical default: the same model, the same geometry up to richness, reached more cheaply and a little more reliably.

Both forms work, and we built each from a particular rotation: one Haar-random dense matrix, one fixed pairing of adjacent coordinates. That two such different choices yield the same model already suggests the specific rotation is not what matters. The next section makes that question explicit, asking whether it matters at all which of the many $90^\circ$ frames the device supplies.

\section{Which $90^\circ$?}
\label{sec:gauge}

Section~\ref{sec:sparse} left a question about the frame the device supplies: among all the $90^\circ$ frames it could pick, does it matter which one? It does not, and the choice can be made deliberately. Three lines of evidence follow, one imposed, one found in the wild, and one prescribed.

The word for a quantity like this is a \emph{gauge}: a degree of freedom that has to take some value for the model to be written down, but whose particular value leaves the function unchanged. There are many rotations that hold the concept phase $90^\circ$ from the read-out, the task gives no reason to prefer one, and a model must nonetheless end up in one of them. What follows asks whether the model cares which, and then whether we can choose on its behalf.

\subsection{Any device, the same model}
\label{sec:gauge-imposed}

We have already built the device from two very different rotations, a dense Haar matrix and a sparse pairing of adjacent coordinates, and both reached baseline quality. Widening that to twenty-five runs across thirteen distinct devices tells the same story (Figure~\ref{fig:gauge}a). The thirteen are eight dense Haar-random $90^\circ$ rotations and five sparse signed permutations, with the remaining runs training-seed replicates of one of them. Across all twenty-five, LAMBADA spans $0.226$ to $0.287$ (mean $0.263$) and BLiMP $0.788$ to $0.810$, bracketing the baseline, with no device standing out. Whichever $90^\circ$ frame the model is handed, it trains to the same model. (Two further dense runs are set aside: one diverged during training to near-chance accuracy, and one was aborted partway through training and never completed. Like any failed seed these are training failures rather than frame effects.)

\subsection{The gauge in the wild}
\label{sec:gauge-wild}

The device imposes a frame, but the freedom it exploits is present with no device at all. Take two baselines trained from different seeds, align them by their read-outs, and compare the frames in which each holds its concepts (Figure~\ref{fig:gauge}b). At every layer the two models' concept frames sit about $90^\circ$ apart, a cross-model angle of $89.7^\circ$ on average, indistinguishable from the $91.4^\circ$ of a random orientation, while within a single model the same measurement has a noise floor near $20^\circ$. The two networks solve the same task, their read-outs aligning at $R^2 = 0.81$, yet they hold their intermediate concepts in near-orthogonal frames. The task fixes the read-out, the $0^\circ$ answer axis, and leaves the $90^\circ$ frame free; two runs of the same recipe simply fall into different, unrelated choices of it.

\subsection{Prescribing the exact frame}
\label{sec:gauge-named}

A gauge that is free is also a gauge that can be prescribed. If the choice of $90^\circ$ frame carries no consequence for what the model computes, then we should be able to prescribe one in advance and have the model adopt it at no cost. We can.

We draw a fixed orthogonal basis $B$ at random, one per run, and add a term to the training objective asking each concept-phase layer to hold its residual second moment diagonal in $B$. The step schedule of Equation~\ref{eq:step} and the boundary device both remain in place, so the concept phase is still required to sit $90^\circ$ off the read-out and the prescription selects among the $90^\circ$ frames rather than competing with the requirement to be off-axis at all.

The form of the term matters more than it appears to, and the obvious versions fail in ways that are easy to mistake for success, so we give it precisely. Let $\tilde{x}_\ell$ be the normalized residual layer $\ell$ reads, and write its second moment in $B$'s coordinates,
\begin{equation}
M_\ell \;=\; \mathbb{E}\!\left[(B^{\top}\tilde{x}_\ell)(B^{\top}\tilde{x}_\ell)^{\top}\right].
\label{eq:presc-M}
\end{equation}
Normalize it \emph{per pair} into a correlation matrix and penalize what is left off the diagonal:
\begin{equation}
P_\ell \;=\; D_\ell\, M_\ell\, D_\ell,
\qquad D_\ell = \operatorname{diag}(M_\ell)^{-1/2},
\qquad
\mathcal{L}_{\mathrm{frame}} \;=\; \sum_{\ell \in \text{concept}} \; \sum_{i \neq j} \big(P_\ell\big)_{ij}^{2}.
\label{eq:presc-loss}
\end{equation}

Three choices in Equation~\ref{eq:presc-loss} are load-bearing.

The first is the per-pair normalization. The natural alternative is a single global ratio, $\lVert M_{\mathrm{offdiag}}\rVert^2 \mathbin{/} \lVert M\rVert^2$, and it is degenerate: a model can drive it to zero by inflating one direction until that direction dominates the denominator, leaving the numerator untouched. What results is a rank-one second moment parked on a single axis of $B$, which reports as near-perfect alignment while naming one direction rather than a frame. Equation~\ref{eq:presc-loss} closes that route by construction. Because $(P_\ell)_{ii} = 1$ for every $i$, no rescaling of any coordinate can reduce the loss, and $(P_\ell)_{ij}$ for $i,j$ away from an inflated direction does not involve that direction's variance at all. A rank-one $M_\ell$ gives $\lvert (P_\ell)_{ij}\rvert = 1$, which is the loss \emph{maximum}.

The second is that the term is applied per layer rather than to the concept phase pooled. A pooled version is satisfied by aligning the layers that are cheapest to align, and produces a profile concentrated in one or two layers that reads, in aggregate, as if the whole phase complied.

The third is the choice of $B$ itself, which must be drawn at random rather than taken to be the canonical basis. A second moment that is diagonal in the canonical axes is also diagonal in every signed permutation of them, so against a device built from coordinate permutations the canonical target is degenerate and the measured contrast is exactly $1$: the model appears to comply without doing anything. Drawing $B$ from the orthogonal group removes the degeneracy.

Two diagnostics separate a real frame from the failure modes above, and we report both. Deflation: remove the leading eigendirections of the residual and re-measure. A prescription that named a frame is unaffected or strengthened; one that inflated a single direction collapses to chance as soon as that direction is removed. And the participation ratio of $\operatorname{diag}(M_\ell)$: a genuine frame spreads across hundreds of coordinates, while the single-axis solution sits near $1$.

\begin{figure}[H]
\centering
\includegraphics[width=0.98\textwidth]{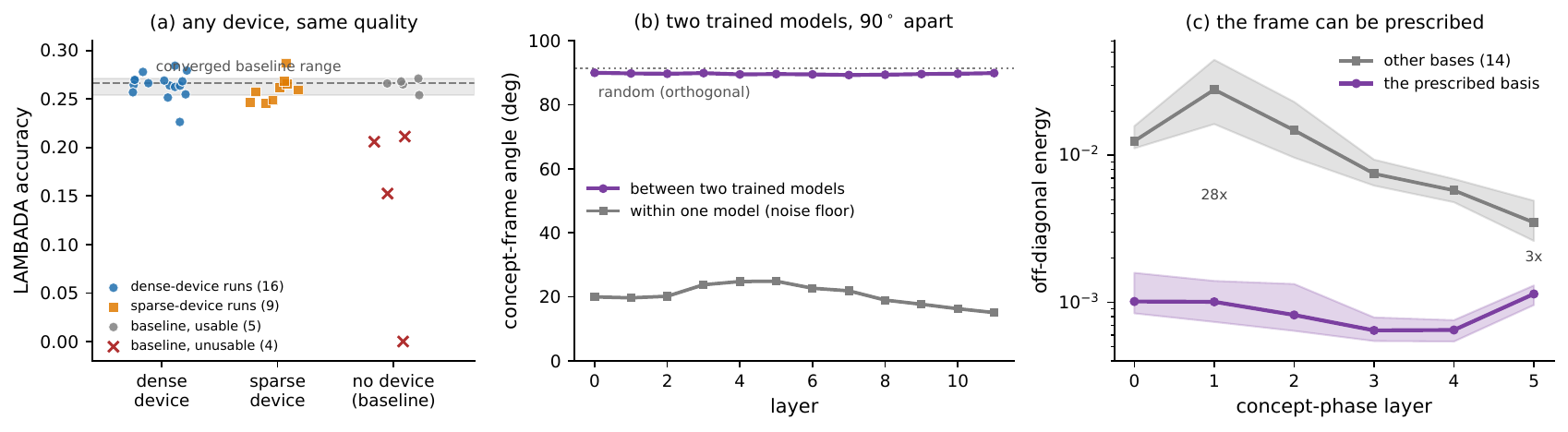}
\caption{The $90^\circ$ frame is free, and it can be prescribed exactly. (a)~Twenty-five device runs, spanning thirteen distinct devices with the remainder training-seed replicates of one of them, set against the nine baseline seeds themselves. Every device run lands in the converged baseline range (band), while ordinary training falls out of it four times in nine (crosses). The comparison is not against a reliably good baseline; it is against one that fails almost half the time. (b)~Two baselines trained from different seeds, aligned by their read-outs, hold their concepts in near-orthogonal frames at every layer (cross-model angle $\approx\!90^\circ$, indistinguishable from the random $91^\circ$, dashed), against a within-model noise floor near $20^\circ$. (c)~Prescribing a randomly drawn basis $B$ during training (seven converged seeds, bands span the runs): the off-diagonal energy of the concept-phase second moment sits $3$ to $28$ times lower in the prescribed basis (purple) than in the comparison bases (grey), at every layer of the concept phase and with no overlap between the two sets, while quality stays inside the baseline range. The task fixes the answer axis and leaves the frame free, and the free frame can be prescribed exactly.}
\label{fig:gauge}
\end{figure}

The models adopt the frame they are given (Figure~\ref{fig:gauge}c). Seven of eight seeds converged, and across them the off-diagonal energy of the concept-phase second moment is $3$ to $28$ times lower in the prescribed basis than in the comparison bases, which include foreign bases drawn the same way and random controls. The separation holds at every one of layers~0--5 rather than concentrating in one or two, and the ranges over runs do not overlap at any depth. It also survives deflation: after removing the top five eigendirections of the residual the prescribed basis remains a factor of four below the others, so the model has aligned a frame rather than parked a single dominant direction on one axis of $B$. Quality is unaffected on the downstream measures, LAMBADA $0.265$ and BLiMP $0.796$, both inside the converged baseline's range, at a perplexity cost of $0.4$ (from $19.1$ to $19.5$). The alignment is a property of the normalized residual the layers are trained through; on the raw residual it is strongest in the first three layers.

Prescribing the frame changes which $90^\circ$ the model uses and nothing else. The angle to the read-out, the participation ratio, and the two-phase profile are all as they were, which is what makes the result a statement about gauge rather than about geometry.

So the $90^\circ$ the device supplies is immaterial to what the model becomes, and it is also ours to specify. The frame is fixed neither by the task nor by the device, the model computes equally well in any of them, and it will compute in whichever one we name. That combination is what makes the off-axis phase a place to put things rather than merely a place things end up: pick any $90^\circ$ you like.

\section{Discussion}
\label{sec:discussion}

\subsection{Supplying a frame increases training reliability}
\label{sec:disc-trainability}

The device was built to install a geometry, and the geometry is what Sections~\ref{sec:device}
and~\ref{sec:sparse} report. The result that was not designed for is convergence. Under the same step
schedule, the same data, and the same budget, two of eight seeds reach a usable model when the
prescription is carried by the loss alone, seven of eight when a dense rotation supplies the frame,
and nine of nine when a sparse one does. The baseline, under no prescription at all, converges five
of nine.

That last comparison crosses conditions and should be read with care, but the direction is hard to
ignore: a fixed, untrained, input-independent rotation inserted at one boundary produced a more
reliable training process than the unconstrained model it was meant to constrain. The device adds no
capacity, and it cannot add quality, because the baseline already reaches the quality the
device-trained models reach. What it removes is a search. A model asked to hold its concept phase off
the read-out has to discover a stable off-axis frame and hold it against the pressure at the same
time, and the cheapest way to satisfy the second demand is to abandon the first, collapsing the
workspace to a few directions. Handing the model the frame separates the two problems and leaves only
the one it is good at.

The general form of this is a claim about gauge freedom. Section~\ref{sec:gauge} shows the choice of
$90^\circ$ frame is immaterial to what the model computes: twenty-five runs across thirteen distinct rotations give the same model,
and two baselines from different seeds land in unrelated frames. A degree of freedom that does not
affect the answer is usually treated as harmless. It is not harmless during optimization. The model
must still pick a value, the picking is done by gradient descent under pressure, and the failures we
observe are failures of that search rather than of capacity. Fixing the gauge in advance costs nothing
at convergence and removes a way to fail.

\subsection{What asking for decodability costs}
\label{sec:disc-decodability}

A growing line of work applies pressure in the opposite direction, training intermediate layers to be
readable through the final unembedding so that decoding can stop early
\citep{elhoushi2024layerskip,schuster2022calm,eetuning2024}. That work reports inference speedups and
does not report what the pressure does to the representation. Section~\ref{sec:forcing} supplies one
measurement of it. A model forced on-axis at every layer reaches parity on quality and pays for it in
geometry: the concept-phase workspace falls from about twenty-five effective dimensions to a mean of
fourteen, with individual runs spanning nine to twenty-one.

Two features of that result bound how far it should be carried. The geometry is distorted, and
substantially so. And the distortion is invisible to the instruments that would normally be used to
look for it, since LAMBADA, BLiMP, and perplexity all sit inside the baseline range. At this scale and
on these benchmarks, the pressure is free.

Whether it stays free is a separate question and this paper does not answer it. A representation of
reduced effective rank may cost nothing on next-token benchmarks at $125$M parameters and cost
something at a larger scale, or on a task that uses the directions that were removed. What we can say
is bounded: the geometry moves, it moves a great deal, and a study measuring only downstream accuracy
would not know that it had. Reporting effective dimensionality alongside the speedup is cheap, and it
is where this particular change shows up.

\subsection{Placing a representation rather than finding one}
\label{sec:disc-placing}

Most of interpretability is archaeology. A model is trained, it arrives at some internal organization
of its own choosing, and the work is to recover that organization after the fact with probes, lenses,
and dictionaries. The organization is treated as a fact about the model to be discovered.

Some of it is not a fact about the model. Section~\ref{sec:gauge} shows that the frame the concept
phase computes in is a free gauge: the task fixes the answer axis and leaves the rest unconstrained,
and two models trained from different seeds land in frames as unrelated as chance allows even though
their read-outs agree at $R^2 = 0.81$. That is worth stating plainly, because it means a substantial
part of what we spend effort recovering was never determined by the data, the objective, or the
architecture. It was determined by the seed.

And what the seed can determine, we can determine instead. Prescribing a basis in advance and asking the
concept phase to hold its second moment diagonal in it produces models that adopt the named frame,
uniformly across the concept phase, with downstream quality unchanged. The cost of choosing is
approximately the cost of not choosing.

Two consequences follow. The first is comparability. Two models are currently hard to compare
internally because they are in unrelated frames, which is a problem the field usually attributes to
representational idiosyncrasy; part of it is simply an unfixed gauge, and models trained into a shared
prescribed basis would be directly comparable at every concept-phase layer without an alignment step. The
second is that legibility becomes a design parameter rather than a discovery target. If the frame is
ours to set, it can be set to something chosen for its interpretability, and the model will compute in
it. That is the more useful reading of a null result about gauge freedom: the fact that the
choice does not matter to the model is exactly what makes it available to us.

\subsection{Limits}
\label{sec:disc-limits}

The models here are a single architecture family at one scale, trained on one corpus, and the phase
structure we prescribe is a strong intervention rather than a gentle one. Whether the same lottery and
the same fix appear at larger scale is open, and the convergence rates we report are over flocks of
eight and nine seeds, which is enough to separate two-of-eight from nine-of-nine but not enough to
resolve smaller differences. The insulation measurement is causal and specific; the account of it, that
attention needs somewhere to average that the unembedding discards, is an inference from that
measurement rather than an independent result.

\section{Related Work}
\label{sec:related}

\paragraph{Reading intermediate states in vocabulary space.}
The logit lens \citep{nostalgebraist2020} decodes a transformer's intermediate
residual states by applying the frozen output unembedding directly, and fails on early and middle
layers. The tuned lens \citep{belrose2023} repairs this by learning a per-layer affine
translator into the final layer's output space, and it reads the model's latent prediction, its
\emph{state}, at each depth far more faithfully than the logit lens. It is our reading instrument,
not a rival.
Its account of why the logit lens fails is a claim about basis, that a hidden state is written in the
basis of its own layer rather than the one the final unembedding expects, and our measurements are
consistent with it. What we add is a claim that account does not make. That a \emph{probe} can undo a
change of basis says nothing about whether the \emph{model} reaches its answer that way, and the two
are easy to conflate. They come apart here: across the commit a best-fit rotation captures
$R^2_{\mathrm{rot}} = 0.38$ and most of the committed state is orthogonal to any rotation of what came
before (Section~\ref{sec:baseline-additive}), so the on-axis answer is written rather than re-based.
No decoder of the state can distinguish an additively built prediction from a rotated one, since the
state is the same either way, which is why this needs the geometry rather than a lens to settle. A parallel
line projects feed-forward value vectors into the vocabulary to interpret their contribution
\citep{geva2021,geva2022}. We use the tuned lens as an instrument rather than propose
one, and our contribution is elsewhere: a functional account (Section~\ref{sec:baseline-offaxis})
of \emph{why} the content sits off the readout axis, which the lens literature documents but does
not explain, and a device that supplies the off-axis frame rather than inferring it after the fact
(Section~\ref{sec:device}).

\paragraph{Iterative refinement and the depth alignment sweep.}
Residual connections have long been characterized as encouraging \emph{iterative inference}:
each block nudges features toward the output across depth \citep{jastrzebski2018residual}. The
most direct measurement of the phenomenon we study is \citet{lys2026causalshift}, who show that
hidden token representations switch from input (current-token) to output (next-token) alignment
deep in the network, and name this the ``causal shift''; \citet{unraveling2025refinement}
likewise trace how per-layer distributions converge toward the final output. A contemporaneous
geometric account, \citet{lombardo2026threephases}, divides next-token prediction into three
depth phases using representation lenses on the Grassmann manifold, and observes, as we
do, that layer updates are approximately orthogonal to the residual stream throughout; but its
similarity measures are layer-relative and invariant to the readout axis (like the
representational-similarity metrics below), so our measurement is complementary rather than
overlapping: we decompose the commit into rotation versus freshly-written content against the
\emph{fixed} unembedding, which a layer-relative similarity cannot express. Crucially,
\citet{lys2026causalshift} frame the off-axis condition as a \emph{structural misalignment}, residual
connections tie activations to the current token while supervision targets the next, and propose
to mitigate it via residual attenuation and gating. Our measurements extend that observation and suggest a
complementary reading. Taking their mitigation to its limit is informative in its own right: pinning
every layer onto the readout axis does train to parity, and it does so by collapsing the off-axis
workspace the model would otherwise use, which is what leads us to read the off-axis condition as
functional rather than as something to be corrected.

\paragraph{Off-readout-axis computation and the unembedding null space.}
The unembedding matrix has a sharp effective-rank drop, leaving a large approximate null space in
the residual stream. \citet{stolfo2024confidence} show that ``entropy neurons'' write almost
exclusively into this null space, modulating residual norm and output confidence with minimal
direct effect on the logits: the closest documented example of a \emph{functional} off-readout
subspace, though for confidence control rather than cross-token composition. That representations
pack many non-orthogonal features into a shared space is the subject of superposition
\citep{elhage2022superposition,bricken2023monosemanticity}. We identify a distinct function for
the off-readout subspace: it \emph{insulates} attention's mixing of concepts across tokens,
letting values compose additively without averaging their vocabulary predictions into a blur.

\paragraph{Linear concept directions and additive editing.}
That high-level concepts are encoded as linear directions is the linear representation hypothesis,
formalized by \citet{park2024linear}, who show its geometry is well-posed only under a particular
(causal) inner product, a choice of frame, which parallels our finding that the concept phase is
legible in the right frame and near-empty in the raw vocabulary one. The activation-editing line
exploits the residual stream's additivity directly: activation addition \citep{turner2023actadd}
and contrastive activation addition \citep{rimsky2024caa} steer a model by \emph{adding} a concept
direction to the residual. That such additions work is evidence the stream is additive and that
concepts lie along directions; we show the model's own forward pass reaches its answer the same
way, writing a fresh on-axis direction late rather than rotating accumulated content onto the
readout, and we localize where and when that write happens.

\paragraph{Representational similarity across depth.}
A mature line measures how similar layer representations are across depth and across models: CKA
\citep{kornblith2019cka}, SVCCA \citep{raghu2017svcca}, PWCCA \citep{morcos2018pwcca}, and the
block-structure analysis of \citet{nguyen2021wide}; recent work reports that concept directions
rotate substantially between where they enter and exit the network \citep{gem2026geometric}, and
that concept formation and separability extend across depth \citep{caz2026concept}. Their invariances differ in ways that matter here: all three are at least invariant to orthogonal
transformation, and the CCA-based measures are additionally invariant to any invertible linear
transformation, which CKA is not \citep{kornblith2019cka}. What none of them can express is rotation
relative to a \emph{fixed} external axis, since each compares one layer's representation to another's
rather than to the read-out; our angle-to-readout is therefore complementary to, rather than subsumed
by, representational-similarity analysis.

\paragraph{The residual-stream gauge and prescribed geometry.}
\citet{elhage2021} establish that the residual stream has \emph{no privileged basis}:
any global rotation can be absorbed into adjacent trainable weights to yield an identical function.
This is what makes our inter-layer rotation a usable intervention: it is functionally vacuous
unless it binds against a non-absorbable, privileged object, and the tied frozen unembedding is
exactly such an object; per-channel normalization gains are another
\citep{transformercircuits2023privileged,sweeney2026signedperm}, and in
practice a privileged basis emerges anyway, attributed to Adam's non-rotation-equivariant updates
\citep{adam2023privilegedbasis}. Among prescribed-geometry architectures, rotary position embedding
\citep{su2021roformer} applies a fixed rotation across \emph{position}; white-box CRATE transformers
\citep{yu2023crate,yu2024crateextended} unroll each layer as an optimization step toward incoherent,
sparse subspaces. Closest to our device, the Three-Phase Transformer \citep{ayyash2026threephase} inserts a Givens
rotation between the attention and feed-forward sublayers, mechanically the same family of
inter-sublayer rotation. Its angles are \emph{learned}, initialized on a depth-linear schedule and
free to move from there, the rotation overwrites the stream rather than adding to it, and it rotates
partitioned channels among themselves as an optimization prior, improving perplexity by $7.2\%$ and
convergence by $1.93\times$ at $123$M. It is never oriented toward the unembedding or toward
legibility. Outside the interpretability literature, structured pruning also rotates
the residual stream per block, aligning it with the activation principal components in order to
choose what to delete \citep{ashkboos2024slicegpt}. Neither is the intervention we make. The
Three-Phase Transformer's angles are learned rather than prescribed, and SliceGPT's frames are read
off a network that has already finished training, so both describe or refine a model rather than
determine one. Ours is fixed in advance and present throughout training, oriented toward the
read-out, and installed so that the phase structure is legible. We have not found that combination
elsewhere. The gauge freedom \citet{elhage2021} identify
is usually invoked to explain why a quantity \emph{cannot} be measured; we use it in the other
direction, showing that the free frame can be named in advance and that the model adopts the one we
choose (Section~\ref{sec:gauge-named}).

\paragraph{Workspaces and off-readout structure.}
Two concurrent 2026 works locate an off-readout structure like the one we describe but
\emph{discover} it post hoc rather than characterize its function:
\citet{transformercircuits2026workspace} identify a verbalizable ``global workspace'' in the residual
stream, and \citet{huang2026subspacepartition} decompose representation space into interpretable
subspaces by unsupervised analysis. Discrete-bottleneck methods such as Codebook Features
\citep{tamkin2023codebook} likewise reconstruct concept structure post hoc; that the
off-axis subspace carries structure at all is consistent with standard sparse dictionary decoding
\citep{bricken2023monosemanticity}. None of these gives the functional account, insulation and the
division of labor between the two phases, which is the substance of this paper, and none of them
supplies the workspace rather than finding it.

\paragraph{The layer-to-layer rotation.}
That the residual basis moves with depth is implicit in the tuned lens, whose per-layer affine
translators exist because the frame does not stay put, and it has recently been measured directly:
\citet{bhattacharya2026residual} decompose each transition by orthogonal Procrustes into a rigid
rotation and a non-rigid remainder, while \citet{fernando2026dynamics} take a full eigendecomposition
of the layer Jacobian and report early layers non-normal and rotation-dominated, late layers
approaching symmetry. That Jacobian is a related but distinct linearization from the best linear
predictor used here, and Section~\ref{sec:baseline-rotation} measures how far the two come apart. \citet{lad2024remarkable} bound from another direction how much any one transition can matter, reporting that deleting or swapping adjacent layers retains $72$--$95\%$ of prediction accuracy. Neither of the geometric accounts reports an angle, and \citet{bhattacharya2026residual} caution that
their own Procrustes magnitudes may sit at the concentration scale $\sqrt{2d}$ rather than measuring
a learned quantity; the rotation we report is three times below that scale and is measured against an
explicit no-motion null (Section~\ref{sec:baseline-additive}). The gap between a nonlinear map's
best linear approximation and its mean Jacobian, which supplies most of the turn
(Section~\ref{sec:baseline-rotation}), is a classical object: the least-squares fit recovers the true
direction only under an elliptical design and departs from it by a bounded amount otherwise
\citep{liduan1989,duanli1991}, and the Gaussian-surrogate linearization of a transformer feed-forward
block has been derived in closed form on that identity \citep{belrose2025polynomials}. What is new
here is measuring that departure inside a trained network and connecting it to the frame the concept
phase computes in.

\paragraph{Deep supervision, early exit, and readout-alignment pressure.}
A complementary line applies the \emph{opposite} pressure, training intermediate layers to be
decodable through a shared readout for early exit and self-speculative decoding: LayerSkip
\citep{elhoushi2024layerskip}, confident adaptive language modeling \citep{schuster2022calm},
EE-Tuning \citep{eetuning2024}, and short-cutting via learned linear maps \citep{yomdin2024jump}.
This line \emph{pushes} intermediate representations toward the readout, and it does report quality:
\citet{schuster2022calm} make a calibrated bound on the quality loss the central contribution, and
\citet{elhoushi2024layerskip} report perplexity together with eleven downstream benchmarks against an
unmodified baseline, including a regression they state plainly. The quantity we add is the effect on the
\emph{representation} itself: forcing every layer onto the read-out reaches parity on quality while
the concept-phase workspace falls from about twenty-five effective dimensions to a mean of fourteen.

\section{Conclusion}
\label{sec:conclusion}

A transformer holds its concepts off the axis it answers on, and it does so for a reason. The
off-axis subspace is where attention can average values across tokens without averaging their
vocabulary predictions into a blur, and the cost of removing that insulation is large and
specific: $64$--$84\times$ a matched random rotation, entirely through the cross-token path. The
answer is not the accumulated content turned onto the read-out. It is written fresh, late, and
added.

Reading that geometry and imposing it are different problems, and the difference is a frame. A loss
that names the target angle asks the model to discover a stable off-axis subspace and hold it under
pressure at the same time, and most seeds fail at the first. One fixed rotation inserted at the phase
boundary removes that burden and the prescription becomes reliable, at baseline quality, with the
workspace intact. Reliable is not a figure of speech: the absorbable frame converged on all nine
seeds we trained, where ordinary unpressured training converged on five of nine. The device buys no
raw quality, and it was never going to: the baseline already reaches that quality on its own. What it
buys is that the geometry can be specified in advance instead of hoped for, and that specifying it
costs nothing in quality while taking trainability from roughly half of runs to all of them.

The reverse pressure is worth recording for the same reason. Pulling every layer onto the read-out,
which is what training for early exit asks for, leaves perplexity, LAMBADA and BLiMP inside the
baseline range while the concept-phase workspace falls from about twenty-five effective dimensions to
fourteen. The geometry moves a great deal and the usual instruments do not report it. Effective
dimensionality is cheap to measure alongside an accuracy number, and it is where this particular
change is visible.

The frame itself turns out to carry remarkably little. A sparse rotation the network can absorb works
as well as a dense one it cannot, and leaves the concept phase half the effective dimensionality. Twenty-five runs across thirteen distinct rotations give the same model. Two baselines trained from different seeds hold their concepts in
frames as unrelated as chance allows. And a basis drawn at random and prescribed before training is
adopted, uniformly across the concept phase, with downstream quality unchanged. The task fixes the
answer axis and leaves the rest free, and that freedom is usable.

What follows from that is a change in what gauge freedom is for. It is usually cited as a reason a
quantity cannot be measured: the residual stream has no privileged basis, so any statement about a
direction is a statement up to a rotation nobody chose. Read the other way, an unfixed gauge is a
control surface. Part of what interpretability currently works to recover after the fact was never
determined by the data, the objective, or the architecture, and a degree of freedom the model is
indifferent to is one we are free to set. A representation we can place is a better starting point
than one we can only find.

\bibliographystyle{plainnat}
\bibliography{refs}
\end{document}